\documentclass[journal]{IEEEtran}

\usepackage{amsmath}
\usepackage{amssymb}
\usepackage{amsthm}

\usepackage{algorithm}
\usepackage{algpseudocode}

\usepackage{graphicx}
\usepackage{booktabs}
\usepackage{multirow}
\usepackage{makecell}

\usepackage{enumitem}
\usepackage{float}
\usepackage{url}
\usepackage{hyperref}
\usepackage{cite}

\usepackage{xcolor}
\usepackage{tcolorbox}
\usepackage{listings}

\tcbuselibrary{breakable}

\tcbset{
    colback=gray!5,
    colframe=black!70,
    boxrule=0.5pt,
    arc=3pt,
    left=6pt,
    right=6pt,
    top=6pt,
    bottom=6pt,
    width=\textwidth,
    breakable
}

\begin{document}

\title{LLM-based Multimodal Personality Recognition via Facial Action Unit-Text Semantic Fusion} 

\author{Tianyi~Zhang$^{1}$,~\IEEEmembership{Senior Member,~IEEE,}  
        Wei~Shan$^{1}$, ~\IEEEmembership{Student Member,~IEEE,}
        Yuan~Zong$^{*}$, ~\IEEEmembership{Member,~IEEE,}
        Tianhua~Qi, ~\IEEEmembership{Student Member,~IEEE,}
        and Wenming Zheng, 
       ~\IEEEmembership{Senior Member,~IEEE,}
\IEEEcompsocitemizethanks{ \IEEEcompsocthanksitem Tianyi Zhang, Wei Shan, Yuan Zong, Tianhua Qi and Wenming Zheng are with the Key Laboratory of Child Development and Learning Science (Ministry of Education), School of Biological Sciences and Medical Engineering, Southeast University, Nanjing, China. Corresponding author: Yuan Zong:{xhzongyuan@seu.edu.cn}}
\thanks{$^1$ both authors contributed equally to this work}}

\markboth{IEEE Transactions on Affective Computing}{IEEE TAFFC}

\maketitle

\begin{abstract}
Personality recognition in asynchronous video interviews (AVIs) has become increasingly important due to their widespread adoption in modern recruitment.  
Existing approaches often rely on large language models (LLMs) to analyze textual responses of interviewees in AVI. 
However, unimodel methods often suffer from information loss (e.g., ignore facial cues).  
In contrast, multimodal methods that employ full-face images or sparsely sampled frames can discard fine-grained temporal dynamics critical for accurate personality assessment.  
To overcome these limitations, we propose an LLM-based framework that semantically fuse facial action units (AUs) with textual responses of AVI.  
AU sequences are first converted into interpretable textual descriptions, which are then fused with participants' textual responses through an LLM.
A lightweight regression head transforms the resulting embeddings into continuous personality scores without disrupting the underlying semantic space.
Experiments on the AVI-6 benchmark demonstrate consistent improvements over most baselines, with lower prediction errors and stronger correlations with human-rated scores across multiple traits.
Further analysis reveals that AU-derived semantic representations offer complementary non-verbal cues to textual responses. 
Decoupling semantic understanding from regression prediction within the LLM also leads to greater training stability and clearer interpretability.
Overall, these findings demonstrate that AU-text fusion provides a psychologically grounded and computationally efficient framework for personality recognition in AVIs.
\end{abstract}



\section{Introduction}
Personality traits reflect distinctive patterns in individuals' thinking, feeling, and behavior \cite{diener2019personality}. 
These traits capture fundamental differences among individuals and show consistency across contexts, as well as stability over time \cite{matthews2003personality}. 
Personality recognition is the task to assess these characteristics from observable cues and has applications in human-centered domains such as personalized product recommendations \cite{10.1016/j.ipm.2022.103256}, video-based recruitment \cite{kaya2017job}, and human-computer interaction \cite{wen2024affectivenli}. 
Among these applications, Asynchronous Video Interviews (AVIs) represent a particularly essential and challenging scenario.
In AVIs, candidates respond to pre-determined questions without live interaction. AVIs offer a flexible way of prescreening and reduce interview costs due to their structured interview format. 
Thus, this type of assessment has increased rapidly over the past few years. According to \textit{QYResearch} \cite{peng2023deposit}, AVI processing volume rose from 26 million to over 40 million between 2022 and 2024 (i.e., a 53.85\% increase).


With the emergence of LLMs, the ability to recognize personality traits from textual input has advanced considerably, as these models demonstrate strong capabilities in capturing latent psychological cues even under zero-shot conditions \cite{ji2023chatgptgoodpersonalityrecognizer}. 
Despite these advances, LLM-driven text-based approaches remain constrained by their reliance on a single modality \cite{Ganesan2023, Yang2023, Zhang2024}. 
For instance, individuals may verbally describe themselves as outgoing or confident, yet exhibit nonverbal cues (e.g., reduced vocal intensity or limited facial expressivity) that convey social inhibition. 
Psychological studies on emotional communication have shown that when verbal and nonverbal signals are inconsistent, observers tend to rely more on nonverbal channels to infer a sender's internal state \cite{DePaulo1979Inconsistent}. 
These findings imply that text-only personality assessment may overlook critical nonverbal discrepancies, thereby limiting its validity and accuracy.
Consistent with this view, \textit{Brunswik's Lens Model} \cite{HirschmüllerSarah2013TDLM} emphasizes that personality traits are accurately perceived through the fusion of multiple cues.
Meta-analytic evidence further supports that multimodal information enables more comprehensive personality assessment \cite{AmbadyNalini1992TSoE}. 
In AVI settings, textual information derived from transcripts is often degraded by noise and affected by the quality of the transcription tools. 
Both psychological theory and empirical evidence thus underscore the limitations of single-modality approaches in capturing the full spectrum of personality-relevant information.

Consequently, researchers have shifted toward multimodal modeling to address the limitations of unimodal approaches by combining visual, auditory, and linguistic cues. 
These methods typically employ modality-specific feature extractors to obtain representations for each modality, followed by either feature-level fusion, which merges features from all modalities into a unified embedding \cite{FeatureLevelFusion}, or decision-level fusion, which aggregates separate predictions through weighting or voting \cite{DecisionLevelFusion}. 
Building on these foundations, subsequent work has introduced attention mechanisms and cross-modal interaction modules to enhance complementarity among modalities and achieve substantial gains in predictive performance \cite{11134060,10386376}.
More recently, graph-based multimodal learning approaches have been explored to model complex inter-modal relationships in structured representation spaces, particularly in affective-related prediction tasks such as student engagement modeling \cite{LI2024102224}. 
Furthermore, multimodal large language models (MLLMs, e.g., GPT-4V \cite{GPT-4V}, LLaVA \cite{LLaVA}, and Qwen-VL \cite{Qwen-VL}) have emerged as a unified paradigm that integrates perception and reasoning within a single framework. 
By coupling modality-specific encoders with a large language model backbone, they combine visual, auditory, and textual information to support robust generalization and end-to-end multimodal understanding across diverse tasks \cite{MLLM}. 

Despite these advances, both traditional multimodal methods and MLLM-based approaches often rely on frame-level preprocessing, which disrupts the inherent temporal continuity of facial behaviors. 
Such disruption undermines the stability and coherence of temporal cues that are essential for personality assessment, since personality traits manifest as consistent behavioral tendencies rather than transient emotional fluctuations. 
When temporal continuity is broken, certain expressive or regulatory features that persist across time are no longer perceptually available to the model. Consequently, this information loss can reduce the ecological validity of personality assessment. 
From a psychological perspective, this loss of temporal availability also contradicts the \textit{Realistic Accuracy Model} (RAM) \cite{RAM}, which posits that accurate personality perception depends on the accessibility of relevant behavioral cues in natural contexts.

In addition to this methodological limitation, the setting of AVIs introduce a practical challenge of their own. 
Participants in AVIs are typically asked to answer questions in a formal interview setting \cite{Brenner_Ortner_Fay_2016}. Thus, their facial movements tend to be restrained and only limited global visual variation for the model to exploit \cite{SieverdingMonika2009'CEc, FeilerAmandaR.2016BEoJ}. 
This low-intensity expressivity further restricts the effectiveness of full-face representations. 
Under such conditions, micro-level cues encoded by facial action units (AUs) offer a more sensitive and psychologically grounded modality for recognizing stable personality signals. 


Motivated by these limitations, we propose an LLM-based framework that fuses AU-derived semantic representations with textual information.
First, we adopt a keyframe-centered small-window strategy to preserve AU-based local temporal variations. 
This strategy mitigates the disruption of temporal continuity caused by uniform frame sampling.
Second, we conduct AU selection to reduce noise and computational burden introduced by irrelevant AUs. 
AU selection is formulated as a multi-objective optimization problem that balances predictive accuracy, selection stability, and feature compactness. 
This problem is solved by using simulated annealing combined with a Pareto-optimal strategy to obtain the most task-relevant AU subset.
After obtaining the optimal AUs set, we align the low-dimensional AU intensity features with a high-dimensional semantic space by leveraging LLM to convert AU sequences into natural language descriptions. 
This allows AU information to be jointly modeled with textual responses in the LLM in a semantically coherent manner.
Finally, we decouple semantic understanding from regression prediction. 
The LLM focuses on modeling multimodal semantic representations, and then a lightweight regression head maps the fused embeddings to continuous personality scores.  
This approach mitigates the limitations of LLMs in text-to-numerical mapping tasks.
These design choices enable the framework to achieve three key contributions in multimodal personality recognition:
\begin{itemize}
\item We propose an LLM-based framework that fuses AU-derived semantic representations with textual information. 
By modeling facial micro-dynamics through keyframe-centered AU windows, our approach captures fine-grained temporal patterns and enables efficient continuous personality assessment. 
\item Our results indicate that the proposed framework consistently outperforms single-modality methods and existing multimodal baselines across multiple regression metrics and personality traits. 
It achieves substantial reductions in error and stronger correlations with human-rated scores. 
\item We show that AU-derived semantic representations provide meaningful non-verbal cues that complement textual responses in asynchronous video interviews. 
Furthermore, we reveal that separating semantic understanding from regression prediction in LLMs stabilizes training, improves predictive accuracy, and preserves interpretability. 
These findings highlight the importance of combining psychologically grounded signals with careful model design for robust personality assessment.
\end{itemize}

\section{Related Work}
This section reviews previous research relevant to this study from three perspectives.
First, we introduce the psychological foundation provided by the HEXACO personality model. 
It provides a validated psychological framework for describing individual differences and acts as a benchmark for evaluation. 
Second, we summarize AU-based approaches for personality recognition, with a focus on their potential as a complementary modality for automated assessment. 
Third, we discuss large language model (LLM)-based personality recognition, covering methods that rely on text alone and approaches that fuse multiple modalities. 

\subsection{The HEXACO Personality Model}
The HEXACO personality model \cite{HEXACO2007} is widely used in psychological research due to its strong reliability and cross-cultural validity \cite{ThalmayerAmberGayle2011CVoB, HEXACOPersonalityEstimation}. 
It defines six personality dimensions: Honesty-Humility (H), Emotionality (E), eXtraversion (X), Agreeableness (A), Conscientiousness (C), and Openness to Experience (O). 
The inclusion of Honesty-Humility extends the traditional Big Five framework to better characterize moral and prosocial tendencies \cite{Hfactor}, which enhances its applicability in diverse behavioral and cultural settings. 
Empirical evidence indicates that H, X, A, and C are closely related to behavioral traits that influence asynchronous video interviews (AVIs), including job performance, impression management, and ethical judgment \cite{ThielmannIsabel2020PaPB, PLETZER2019369}. 
Therefore, HEXACO provides both the theoretical foundation and the evaluation criteria in this study.

\subsection{Facial behavior-based personality recognition}
AU-based representations provide psychologically interpretable indicators of facial muscle activations. 
Their established relevance to interpersonal perception makes them a compelling modality for personality assessment. 
Gavrilescu and Vizireanu \cite{AU16PF} introduced a three-layer neural network that extracts frame-level AU intensities and constructs AU activity maps to predict the Sixteen Personalities. 
The model achieves over 80\% accuracy on several traits and demonstrates significant associations between high-intensity AUs and personality-related behaviors. 
Cai and Liu \cite{Cai2022} developed a machine learning approach to predict Big Five personality traits from facial movements in ordinary videos by analyzing 70 facial key points. 
The model used CatBoost regression and produced moderate correlations with questionnaire scores as well as high reliability. 
These findings support the feasibility of automatic personality assessment from facial cues. 
Raj and Mandal \cite{RajK.Sibin2025FAUa} further investigated the association between AUs and Big Five traits during dyadic interactions. 
Their study employed OpenFace 2.0 to capture and calibrate FAUs. 
Correlation analysis revealed at least seven FAUs with significant links to personality dimensions. 
This evidence strengthens the potential for recognizing personality from facial behavioral patterns.
These approaches primarily rely on static or aggregated facial features and often neglect the temporal dynamics of facial behavior.

To address this limitation, recent work has attempted to move beyond static or frame-level representations by modeling temporal facial dynamics. 
For instance, Song et al. \cite{Song2023} proposed a self-supervised framework to learn person-specific facial dynamics by exploiting the natural temporal evolution of facial actions. 
Their approach captures long-term behavioral patterns and encodes identity-specific dynamics into compact representations, demonstrating improved personality prediction performance. 
This line of work highlights the importance of temporal modeling and individual differences in facial behavior.

Despite these advances, AU-based systems face two major limitations for AVI-based personality assessment. 
First, prior AU features even when enhanced with temporal modeling, often appear as raw numerical signals without psychologically meaningful interpretations or behavioral descriptors. 
This abstraction reduces interpretability and limits the utility of AU information in downstream reasoning tasks. 
Second, using AUs as the uni-modal signal for personality assessment can result incomplete signals in AVI scenarios. 
In AVIs, interviewees respond to personality-inducing questions which are designed to activate the personality traits according to \textit{Trait activation theory} \cite{TraitActivationModel}. 
Thus, the most informative cues for each trait predominantly appear in textual responses. 
Relaying on AUs only cannot capture the breadth of personality expressions elicited in this setting.

Consequently, advancing AVI-based personality recognition calls for multimodal approaches that combine AUs with textual responses. 
By grounding AU signals in interpretable semantics and complementing them with trait-inducing textual cues, such frameworks can overcome the inherent limitations of uni-modal AU-based methods and better capture the full spectrum of personality expressions.

\subsection{LLM-based Personality Recognition}
LLM-based Personality Recognition can be categorized into two types: 1) text-based LLMs and 2) multi-modal based LLMs. Since language serves as a natural medium to express personality \cite{Pennebaker19991296}, text-based LLMs such as GPT \cite{gpt}, T5 \cite{t5}, and Gemma \cite{gemma}, which are pretrained on vast corpora and subsequently instruction-tuned, are capable of performing zero-shot and few-shot personality assessment.
Recent studies demonstrate that LLMs effectively extract latent personality cues.  
For instance, Ganesan et al. \cite{Ganesan2023} incorporated psychological questionnaire items into prompts to improve social media personality assessment.  
Ji et al. \cite{ji2023chatgptgoodpersonalityrecognizer} applied chain-of-thought prompting to enhance reasoning consistency. Yang et al. \cite{Yang2023} proposed PsyCoT, a multi-step reasoning framework integrating psychological theory. More recently, Zhang et al. \cite{Zhang2024} evaluated the use of large language models (LLMs) such as GPT-3.5 and GPT-4 for automatic personality assessment from asynchronous video interviews. 
Their results demonstrate comparable or superior zero-shot validity to task-specific AI models and provide interpretable verbal explanations aligned with psychological constructs.  

AVI responses tend to be brief, structured, and lack interactive context \cite{Brenner_Ortner_Fay_2016}, which constrains the richness of linguistic cues available for personality assessment.  
While textual LLM approaches can capture latent trait information, these limitations motivate the fusion of complementary multimodal evidence. To address this issue, multimodal deep learning approaches leverage visual, auditory, and textual cues to capture richer personality signals.
Sun and Zhang \cite{Sun2023} employed co-attention transformers to capture interactions across modalities and improved overall performance. 
Bao et al. \cite{Bao2024} addressed modality heterogeneity through adaptive fusion strategies and specialized regression objectives. 
Duan et al. \cite{Duan2024} fused hierarchical vision transformers with audio and temporal modeling to make full use of the spatiotemporal information. 

In the AVI context, Lv et al. \cite{MWCF2023} proposed a Multi-modal Window-Consistency Fusion (MWCF) network for automatic assessment of soft skills and personality from asynchronous video interviews. 
The network captures short-term consistency across modalities, assigns weights to verbal cues, and incorporates interviewer knowledge through topic-guided keyword representations to improve interpretability. 
Experiments on a real-world interview dataset demonstrated superior prediction performance and highlighted the model's practical potential for informed hiring decisions.

Although these works demonstrate the promise of multimodal learning, many approaches still rely on global visual representations, which omit fine-grained AU-level dynamics with established psychological meaning. 
Furthermore, personality is often treated as a conventional feature-learning task with limited grounding in validated psychological constructs. 
These limitations motivate the need to model micro-temporal facial behaviors and semantical cues for accurate personality assessment.

\section{Methodology}
\label{sec:Methodology}
In this section, we present the proposed personality recognition framework, which enhances the fusion of AU and text information in AVI scenarios. 
The framework consists of four main components: 
(1) \textbf{Preprocessing}: the keyframes are extracted and small temporal windows centered on these keyframes are constructed.
(2) \textbf{Facial action unit (AU) selection}: the most informative AUs are identified using a simulated annealing-based optimization. 
(3) \textbf{AU-derived semantic description generation}: this module converts the selected AU sequences into semantically rich descriptions within each small window.  The semantical descriptions are also integrated into a single summary text representing dynamic facial changes across the entire video.
(4) \textbf{Personality trait prediction}: the main LLM is LoRA-fine-tuned together with a regression head. The LLM's final hidden representations from AU-derived semantic descriptions and textual responses are used to predict personality scores.

\begin{figure*}[htbp]
    \centering
    \includegraphics[width=\linewidth]{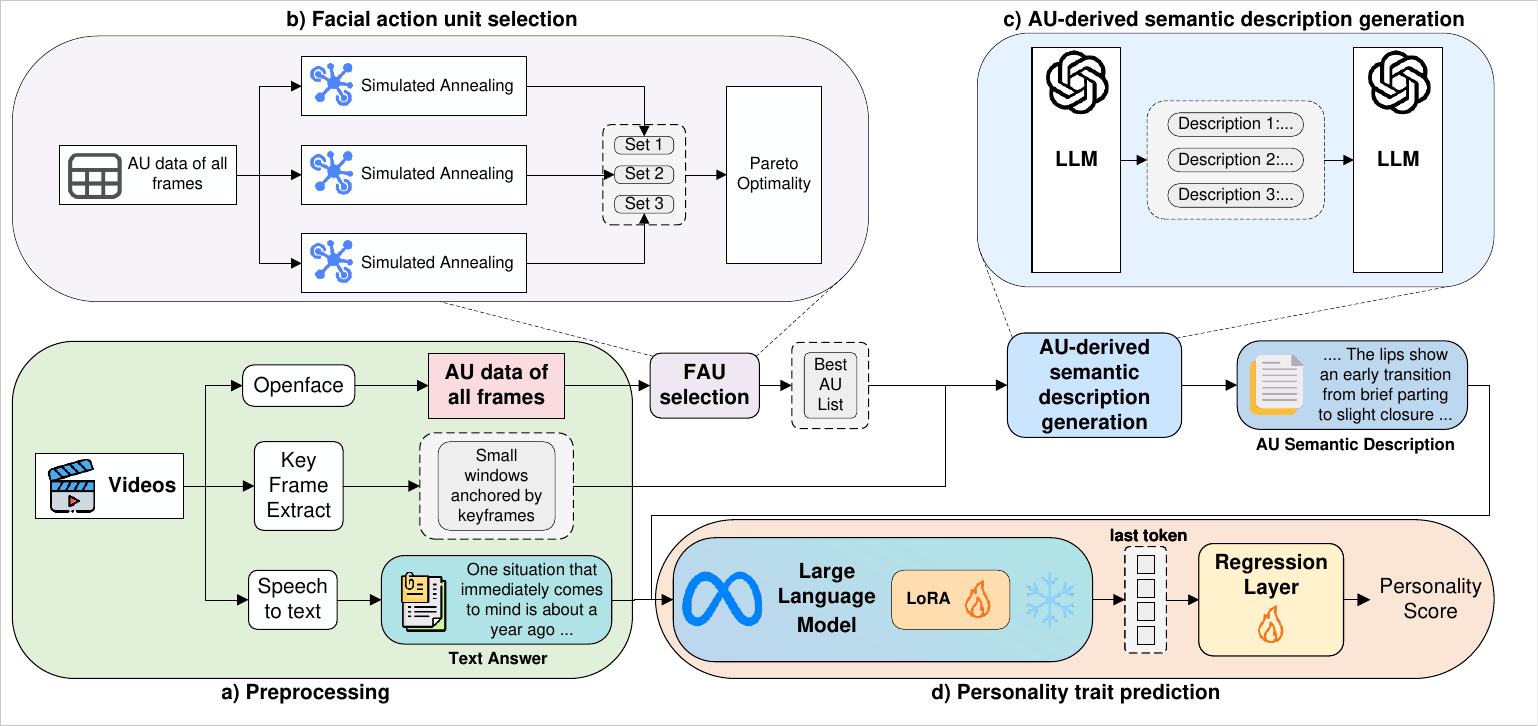}
    \caption{Overview of the proposed multimodal personality assessment framework, 
    consisting of four components: 
    (a) Preprocessing, (b) Facial action unit selection, 
    (c) AU-derived semantic description generation, and 
    (d) Personality trait prediction.}
    \label{fig:framework}
\end{figure*}

Figure~\ref{fig:framework} illustrates the entire processing pipeline from raw video input to continuous personality score prediction.

\subsection{Preprocessing}
Facial Action Units (AUs), defined in the Facial Action Coding System (FACS) \cite{FACS}, provide a fine-grained representation of facial muscle movements.  
We leverage OpenFace \cite{OpenFace} to extract frame-level AU intensity values from interview videos.  
Given the inherently temporal nature of expressive behaviors, we adopt the dynamic prediction mode to enable subject-specific calibration and enhance temporal consistency across frames.
The extracted AU sequences constitute the non-verbal modality input of our framework.

Due to the long durations of the interview videos, the AU intensity sequences extracted by OpenFace span extended temporal ranges and exhibit high temporal variability.  
Directly processing the entire sequence results in considerable computational and memory overhead, which hinders efficient modeling.  
Although frame-sampling strategies reduce the sequence length, they disrupt temporal continuity. 
To maintain localized temporal context while controlling computational cost, we adopt a keyframe-based windowing strategy.  
Prior work has shown that constructing short temporal windows can efficiently model temporal facial dynamics \cite{Song2023}.
Inspired by this, we extract localized AU segments centered on keyframes to model fine-grained temporal variations.

Keyframes are identified by detecting local maxima in a 1D temporal signal of inter-frame motion intensity. 
Specifically, each frame is converted to grayscale, and the mean absolute pixel difference between consecutive frames is computed to obtain a scalar motion value per frame. 
The resulting temporal sequence is smoothed via convolution with a Hanning window to reduce noise. 
Local maxima are then detected using a strict comparison criterion, where a point is considered a peak only if it is strictly greater than its neighboring values. 
No additional thresholding or minimum distance constraint is applied, allowing all motion peaks to be retained.

Since the motion signal is computed from inter-frame differences, each detected peak corresponds to the transition between two frames. 
Therefore, the keyframe is defined as the preceding frame of each detected peak.

To improve computational efficiency, keyframe detection is performed on a temporally subsampled sequence. 
If no local maxima are detected, no keyframe is extracted and the visual modality for that video is represented as an empty input.  

A short temporal window of seven consecutive frames (three preceding and three following the keyframe) is constructed around each selected keyframe.
This window size provides sufficient local temporal context for capturing short-term facial dynamics while avoiding the inclusion of multiple facial events within a single segment.  
A detailed analysis of the window size is provided in Section~\ref{sec:Results}.  
Although keyframe detection is performed on a subsampled sequence for efficiency, constructing windows on the original frame timeline preserves fine-grained temporal dynamics of facial actions.   
This design focuses computation on regions with meaningful AU variation and avoids redundant processing of temporally smooth segments.

\subsection{Facial action unit selection}
AUs provide quantitative measures of facial muscle movements. 
However, their contributions to personality-related behavioral modeling vary among different AUs \cite{RajK.Sibin2025FAUa, 10428080}. 
Using the full set of AUs may introduce redundancy and irrelevant signals, which can negatively affect both computational efficiency and the quality of temporally structured behavioral representations used for personality recognition.

It is important to note that this selection process is not intended to directly optimize the final semantic modeling objective. 
Instead, it learns a structured intermediate selection that preserves temporally informative facial cues relevant to the downstream task.

To operationalize this idea, we introduce a two-stage selection framework. 
In the first stage, a single-run simulated annealing procedure explores candidate AU subsets with an LSTM-based evaluation function defined on AU intensity sequences. 
In the second stage, we extend this process to a multi-objective, multi-run optimization framework that aggregates solutions from multiple simulated annealing runs to construct a Pareto-optimal set.
This formulation jointly considers predictive informativeness, subset diversity (stability across runs), and compactness.

The selected AU subsets are then used to generate semantic descriptions. 
This design provides a compact and structured intermediate representation that bridges low-level facial signals and high-level language-based modeling.

\subsubsection{Single-run Simulated Annealing}
AU selection constitutes a combinatorial optimization problem.
Since each AU may be included or excluded, a search space of size $2^{17}$ is formed. 
Due to the exponential search space, exhaustive enumeration is computationally infeasible. 
Therefore, we adopt simulated annealing (SA), a heuristic optimization strategy.
Its probabilistic search helps avoid convergence to local optima.

\textit{Brunswik's Lens Model} \cite{HirschmüllerSarah2013TDLM} posits that personality recognition is a multi-cue and temporally dependent perceptual process. 
Collaborative activations of multiple AUs provide psychologically meaningful signals, whereas individual AUs in isolation may offer insufficient or ambiguous cues. 
Moreover, personality traits are manifested through dynamic facial expressions rather than static snapshots.
In this context, facial action units (AUs) exhibit both combinatorial dependencies and temporal dynamics.
To incorporate these properties, we introduce a lightweight LSTM-based objective as a structure-aware signal in the search process. 
The goal is to provide a computationally efficient mechanism for assessing whether a candidate AU subset preserves learnable temporal structure in the input sequence space.

Formally, the energy of a candidate subset $S$ is defined as the mean squared error (MSE) of an LSTM trained on the corresponding AU sequences :
\begin{equation}
    E(S) = \text{MSE}_{\text{LSTM}}(S).
\label{equ:energy_function}
\end{equation}
We emphasize that this objective is not designed to approximate downstream personality prediction performance. 
Instead, it is used as a surrogate signal to guide the search toward AU subsets that maintain temporally coherent patterns in the input feature space. 
This design enables efficient combinatorial exploration without full downstream evaluation.

During SA, candidate subsets are generated by randomly toggling the inclusion of a single AU.  
The full AU intensity sequences specified by the candidate subset are fed into the LSTM to compute $E(S)$.  
A candidate subset is accepted if it reduces the energy, or otherwise with probability $\exp(-\Delta E / T)$, where the temperature $T$ gradually decreases according to a predefined cooling schedule until the stopping criterion is satisfied.
This process enables efficient exploration of the AU subset space while relying only on a lightweight temporal model, avoiding repeated evaluation of the full downstream multimodal pipeline.

This evaluation provides a computationally efficient mechanism for guiding subset search toward temporally structured facial representations. It does not assume direct alignment with downstream prediction accuracy, but serves as a structural constraint that reduces search space complexity in the combinatorial optimization process.
The complete SA-based AU selection procedure is summarized in Algorithm~\ref{alg:sa}.
\begin{algorithm}[htbp]
\small
\caption{Simulated Annealing with LSTM-based Evaluation} 
\label{alg:sa}
\begin{algorithmic}[1]
\Require
  $x_0$: initial AUs subset (all AUs); \\
  $T_0$: initial temperature; \\
  $T_{min}$: minimum temperature threshold; \\
  $\alpha$: temperature decay rate;
\Ensure
   Optimal or near-optimal AUs subset;
\State $x_{best} \gets x_0$
\State Train LSTM on training set using $x_{best}$ and compute $E(x_{best})$
\State $T \gets T_0$
\While {$T > T_{min}$}
  \For{$i = 1$ to $n$}
    \State $x_{new} \gets$ Perturb($x_{best}$)
    \State Train LSTM on $x_{new}$ and compute $E(x_{new})$
    \State $\Delta E \gets E(x_{new}) - E(x_{best})$
    \If {$\Delta E < 0$}
      \State $x_{best} \gets x_{new}$
    \Else
      \State Accept $x_{new}$ with probability $\exp(-\Delta E / T)$
    \EndIf
  \EndFor
  \State $T \gets T \times \alpha$
\EndWhile
\end{algorithmic}
\end{algorithm}

\subsubsection{Multi-objective Optimization}
The outcome of a single annealing run depends strongly on random initialization, often resulting in high variance across runs. 
Moreover, optimizing a single surrogate objective, such as the LSTM-based energy, may bias the search toward subsets that are favorable under the surrogate objective but not necessarily optimal for downstream semantic modeling.
To address these limitations, we adopt a multi-objective framework and perform multiple annealing runs using diverse random seeds and initialization strategies.

AU selection for downstream personality recognition involves three conflicting objectives:  
(1) \textit{Temporal Coherence}--the subset should preserve stable and learnable temporal behavioral patterns;  
(2) \textit{Stability}--the subset should be consistent across multiple runs;  
(3) \textit{Compactness}--the number of selected AUs should be minimized to reduce redundancy and improve semantic interpretability.  
These objectives are inherently in tension: improving one may lead to degradation in others.
To balance these objectives, we formulate AU selection as a multi-objective optimization problem, and then construct a Pareto frontier from candidate solutions obtained across multiple runs.
We note that this formulation does not assume that all objectives are equally predictive of downstream performance, but instead provides a structured way to explore trade-offs among complementary properties of AU subsets.

Temporal coherence is quantified using the LSTM-based energy defined in Eq.~(\ref{equ:energy_function}). 
It serves as a surrogate signal reflecting temporal consistency in AU intensity sequences.
Stability is quantified via the Jaccard similarity between AUs subsets:  
\begin{align}
J(A, B) &= \frac{|A \cap B|}{|A \cup B|},\\
Difference(x_i) &= \frac{1}{\sum_{j \neq i} J(x_i, x_j)}.
\label{equ:Difference}
\end{align}
where $x_i$ and $x_j$ denote subsets from different runs. 
Lower difference scores indicate higher consistency across runs. 
Compactness is measured by the number of selected AUs, with smaller subsets preferred.

To achieve balanced trade-offs, we construct the Pareto frontier of non-dominated solutions. 
In order to explain the algorithm principle more intuitively, we give the mathematical definitions as follows:

\noindent A solution $x_1$ \emph{dominates} $x_2$ if
\begin{equation}
f_i(x_1) \le f_i(x_2), \ \forall i, \quad \text{and} \quad \exists j \text{ s.t. } f_j(x_1) < f_j(x_2).
\label{equ:dominate}
\end{equation}
The Pareto frontier is defined as
\begin{equation}
P^* = \{ x^* \in X \mid \nexists x \in X, \, x \prec x^* \}.
\label{equ:pareto}
\end{equation}

To select a representative solution, we use a Utopia-based selection strategy.   
Each objective is normalized as
\begin{equation}
\hat{x}_i = \frac{x_i - \min(P^*)}{\max(P^*) - \min(P^*)}, \quad i = 1, \dots, m,
\label{equ:normalization}
\end{equation}
and the Euclidean distance to the ideal point is computed:
\begin{equation}
d(x) = \left\| \hat{x} \right\|_2, \quad
x_{\text{utopia}} = \arg \min_{x \in P^*} d(x).
\label{equ:distance}
\end{equation}
This subset achieves the most balanced trade-off and is used for generating AU-derived semantic descriptions.
The complete multi-objective optimization framework is summarized in Algorithm~\ref{alg:multi-sa}.
\begin{algorithm}[htbp]
\small
\caption{Multi-objective Multi-run Simulated Annealing Framework} 
\label{alg:multi-sa}
\begin{algorithmic}[1]
\Require
  $Seeds$: random seeds; \\
  $InitStrategies$: initialization strategies; \\
  $SA$: single-run simulated annealing procedure;
\Ensure
  Final optimal AUs subset and Pareto frontier;
\State Initialize result storage
\For{each $s \in Seeds$}
  \For{each $init \in InitStrategies$}
    \State Generate initial AUs subset $x_0$ using $init$
    \State $(x^{*}, MSE) \gets SA(x_0, s)$
    \State Store $(x^{*}, MSE)$
  \EndFor
\EndFor
\State Compute $(MSE, Size, Difference)$ for all candidates
\State Apply Pareto optimization to identify non-dominated solutions
\State Select Utopia solution from the Pareto frontier
\State \Return Final optimal AUs subset and Pareto set
\end{algorithmic}
\end{algorithm}
Only AUs selected through this multi-objective process are used for semantic description generation, ensuring that the resulting representations are temporally coherent, stable, and compact behavioral signals suitable for downstream language-based modeling.

\subsection{AU-derived Semantic Description Generation}
AUs provide quantitative measures of facial muscle movements.  
However, raw AU intensity sequences are low-level and do not readily reveal higher-level patterns associated with personality traits.  
To address this limitation, we introduce a two-step semantic transformation that converts AU temporal sequences into interpretable natural language descriptions.  
This approach allows the model to reason over semantically rich representations instead of purely numerical sequences, which also facilitates effective multimodal fusion in subsequent stages.

\subsubsection{semantic descriptions within small windows generation}
To bridge the gap between low-level behavioral dynamics and high-level semantics, we leverage the comprehension capabilities of large language models (LLMs).  
For each 7-frame window centered on a keyframe, temporal patterns of the selected AUs are embedded into a structured prompt.  
The LLM then generates a concise natural-language description summarizing the facial movements within that interval.  
This process converts numerical AU sequences into semantically rich, human-readable text, enhancing both interpretability and compatibility with downstream LLM-based processing.

To enable consistent and reproducible prompt construction across different stages, we adopt a unified modular prompting framework. 
Each prompt is defined as an ordered composition of the following functional modules: 
(1) role definition: The module specifies the expertise or perspective of the model; 
(2) task description: The module defines the objective of the current stage; 
(3) structured input: The module provides task-relevant information such as AU-derived semantic descriptions, contextual cues, or multimodal signals; 
(4) instruction constraints: The module regulates properties of the output (e.g., semantic coherence, non-redundancy, or psychologically grounded reasoning); and 
(5) optional demonstration examples: The module illustrates the desired output style and level of abstraction. 

Different prompts instantiate different subsets and organizations of these modules depending on the task requirements. 
This modular design enables consistent control of model behavior across stages while adapting to task-specific needs. 

Following the modular prompt design described above, this task-specific prompt additionally incorporates an explicit AU semantic definition module.
It guides the LLM progressively from contextual grounding to AU-based motion interpretation to ensure reliable semantic generation.
\begin{figure}[htbp]
    \centering
    \includegraphics[width=0.95\linewidth]{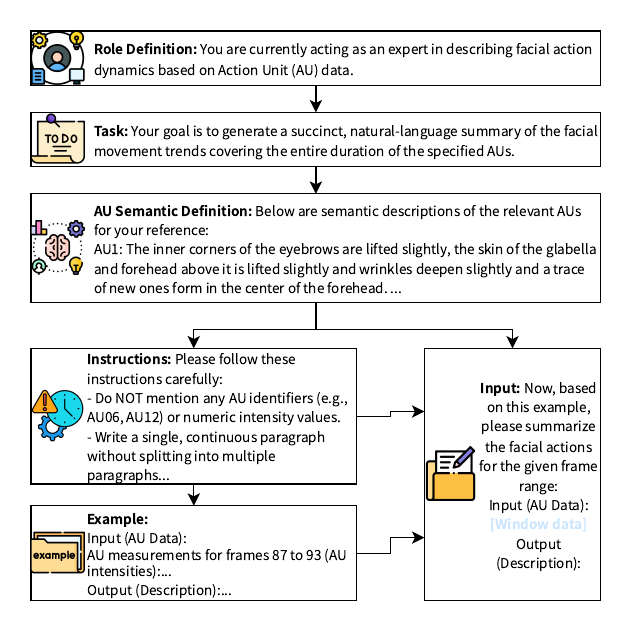}
    \caption{Structured prompt design for generating semantic descriptions of AU sequences within small temporal window}
    \label{fig:prompt1}
\end{figure}
Figure~\ref{fig:prompt1} illustrates a specific instantiation for AU semantic generation, here we highlight two design considerations essential for scientific rigor. 
First, the AU semantic reference set is constructed as a unified and standardized semantic space to ensure consistent interpretability across all AUs.
Among the 17 AUs used, 14 are adopted from established semantic definitions in \cite{AUSemanticDescription}, which also serve as stylistic and structural references for the remaining AUs.
To ensure consistency, the remaining AUs (AU5, AU20, AU45) are generated through constrained in-context prompting. 
Specifically, FACS documentation is used as the semantic grounding source to provide authoritative descriptions of the underlying facial muscle movements and the 14 existing AU definitions are used as in-context exemplars to guide alignment in semantic granularity, vocabulary usage, and descriptive structure.
All generated definitions are subsequently manually reviewed and refined to ensure consistency in descriptive scope, linguistic style, and level of detail.
This process results in a unified AU semantic reference set with consistent granularity and linguistic style.
Second, the demonstration example in the prompt was not handcrafted; instead, it was generated automatically by prompting GPT-4o with a randomly sampled training instance and iteratively refined to ensure that it faithfully reflected the selected AU dynamics while containing no explicit AU labels or numeric values. 
This structured design ensures semantic fidelity and stylistic consistency and effectively links quantitative facial data to high-level behavioral representations.

\subsubsection{global summary descriptions generation}
Directly concatenating descriptions from multiple windows can introduce redundancy, as consistent AU patterns may appear across windows.  
Repeating these sentences increases memory usage and dilutes the proportion of information relevant for personality assessment.

To address these issues, we propose an iterative semantic summarization procedure that consolidates small-window descriptions into a compact global representation.  
The procedure sequentially integrates local descriptions, with the LLM synthesizing content to emphasize consistent and meaningful facial behavior patterns while discarding redundant or invariant details.  
Merging proceeds until all small-window descriptions are incorporated.  
This strategy reduces input length, lowers memory consumption, and yields information-dense representations suitable for downstream multimodal personality recognition.  
The detailed merging procedure is presented in Algorithm~\ref{alg:merge}.
\begin{algorithm}[htbp]
\small
\caption{Merge Small-Window Descriptions into Global Summary}
\label{alg:merge}
\begin{algorithmic}[1]
\Require Text including all small-window descriptions (one per paragraph), personality trait $q$
\Ensure Global summary of the video
\State Read all non-empty lines from the input text into list $L$
\State $merged \gets$ empty string
\For{each $desc$ in $L$}
    \If{$merged$ is empty}
        \State $merged \gets desc$
    \Else
        \State $prompt \gets$ construct from $merged$, $desc$, $q$
        \State $merged \gets \text{LLM}(prompt)$
    \EndIf
\EndFor
\State \textbf{return} $merged$
\end{algorithmic}
\end{algorithm}

Following the same modular prompt design, we construct a task-specific prompt to generate temporally coherent summaries over extended video intervals. 
This prompt guides the LLM in integrating multiple short-window AU descriptions into a single summary.
\begin{figure}[htbp]
    \centering
    \includegraphics[width=0.95\linewidth]{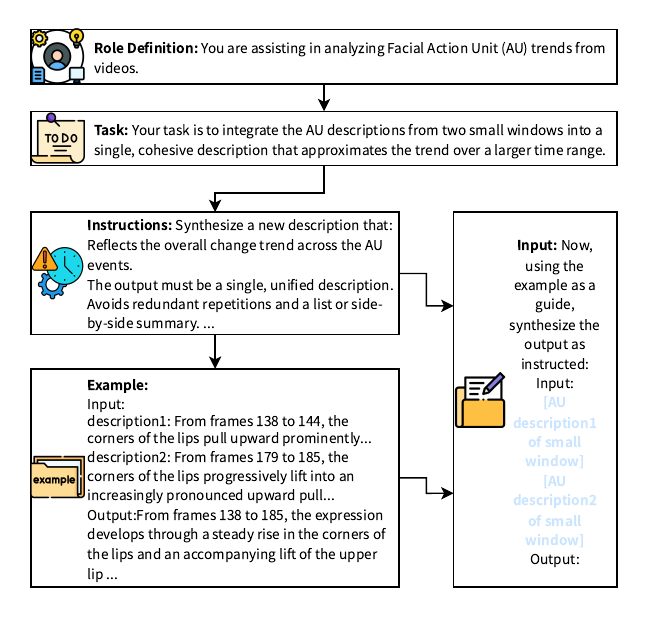}
    \caption{Structured prompt design for generating global summary descriptions}
    \label{fig:prompt2}
\end{figure}
As illustrated in Fig.~\ref{fig:prompt2}, the prompt ensures that the LLM produces a semantically rich, temporally coherent global summary. 
The accompanying instructions enforce avoidance of redundancy while ensuring smooth transitions even when intermediate frames are unavailable. 
A guiding example is curated with the same generation-refinement strategy used for the AU-semantic prompt. 
It specifies the expected level of abstraction and style, and it promotes consistent and controlled integration. 
This design preserves semantic fidelity and temporal continuity in the global descriptions. 
It also enhances their suitability for subsequent fusion with textual interview responses in personality assessment.

\subsection{Personality Trait Prediction}
Although large language models (LLMs) excel in semantic understanding and reasoning, they face inherent limitations in representing continuous numerical values.
Because tokenization cannot encode fine-grained distinctions between floating-point numbers, LLMs often underperform when applied directly to text-to-text regression tasks \cite{understandingllmembeddingsregression}.
Consequently, subtle quantitative variations required for personality score prediction may be lost.

To address this limitation, we decouple semantic understanding from numerical estimation.
Specifically, the LLM generates high-level semantic embeddings from AU descriptions and textual responses.
These embeddings compactly encode multimodal and contextually relevant cues into a semantically coherent representation space.

Under this modular prompt design, we introduce context and trait definition modules to support psychologically grounded reasoning.
The input portion of the prompt provides a recording context to situate behavioral responses, presents AU-derived semantic descriptions alongside textual responses as complementary cues, and defines the target HEXACO trait to anchor the personality trait based on the HEXACO-PI-R trait definitions \cite{HEXACO2007}. 
These inputs ensure that the model has sufficient background for psychologically grounded reasoning and semantic integration.

In some settings, such as encoder-only architectures, the instruction portion (comprising role definition and task description) can be omitted without altering the format of the input AU-derived semantic descriptions or textual responses. 
When included, it frames the LLM as an expert in personality psychology and directs it to generate condensed, trait-informed embeddings that capture personality-relevant behavioral and linguistic patterns.

\begin{figure}[htbp]
    \centering
    \includegraphics[width=0.95\linewidth]{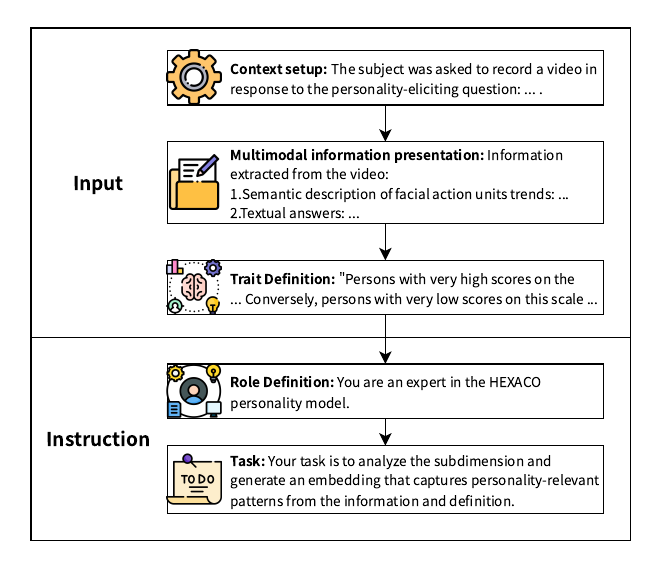}
    \caption{Structured prompt design for generating embeddings for personality trait prediction}
    \label{fig:prompt3}
\end{figure}

This modular design allows consistent interpretation of multimodal cues within a psychologically grounded framework to produce semantically coherent embeddings suitable for downstream regression tasks. 
The hierarchical structure of the prompt is illustrated in Fig.~\ref{fig:prompt3}.

From the processed prompt, the last token's hidden state of the final transformer layer serves as a compact semantic embedding integrating multimodal cues and trait definitions. 
A lightweight regression head then maps this embedding to a continuous personality score. 
The regression head consists of a single linear layer: 
\begin{equation}
\hat{y} = W h + b,
\label{equ:regression_head}
\end{equation}
where \(h \in \mathbb{R}^{4096}\) is the extracted embedding, \(W \in \mathbb{R}^{1 \times 4096}\) and \(b \in \mathbb{R}\) are learnable parameters. 
No normalization is applied to the target labels, which range from 1 to 5, allowing the regression output to directly correspond to the original score scale. 

Limited data and GPU memory make full fine-tuning of the LLM infeasible. 
To enable the LLM to learn personality-relevant knowledge and optimize its semantic representations, we employ LoRA for partial fine-tuning, updating only the low-rank adaptation matrices in the attention projection layers while keeping the original pretrained weights frozen.
Specifically, for each weight matrix $W \in \mathbb{R}^{d \times k}$ in the attention layers, LoRA introduces a low-rank decomposition:
\begin{equation}
\Delta W = BA, \quad B \in \mathbb{R}^{d \times r}, \ A \in \mathbb{R}^{r \times k}, \ r \ll \min(d,k),
\label{equ:lora}
\end{equation}
where $B$ and $A$ are trainable and $r$ is the rank hyperparameter controlling the number of additional parameters. The effective weight used during forward propagation becomes
\begin{equation}
W' = W + \Delta W = W + BA.
\label{equ:lora_forward}
\end{equation}
This approach allows efficient adaptation of the pretrained model with significantly fewer trainable parameters, while preserving the general linguistic knowledge encoded in $W$.  
The regression head parameters are trained jointly with LoRA updates in an end-to-end manner, so that the complete model directly predicts continuous personality scores.  
During training, the input consists of semantically aligned AU descriptions and textual responses, and the model is optimized to minimize the regression loss.  

This design offers several advantages:
\begin{enumerate}[label=\arabic*)]
    \item \textbf{Parameter efficiency:} Only $B$ and $A$ matrices are updated for the LLM. This method reduces memory and computational overhead.
    \item \textbf{Decoupling of semantic and numeric processing:} The LLM learns semantic representations while the regression head focuses on numeric trait estimation.
    \item \textbf{End-to-end adaptability:} The framework allows joint optimization of AU-text embeddings and personality assessment without requiring separate pretraining or post-hoc mapping.
\end{enumerate}

\section{Experiments}
\subsection{Dataset}  
To evaluate the effectiveness of our proposed method, we conduct experiments on a dataset specifically designed for the AVI scenario-AVI-6 \cite{zhang2025assessing}. 
The AVI-6 dataset contains video interview recordings from 644 participants. 
Each participant was asked to answer six questions. 
Among these questions, two are general interview questions commonly used in various assessment contexts, while the remaining four are personality-eliciting questions specifically designed to assess traits of the HEXACO personality model: Honesty-Humility (H), eXtraversion (X), Agreeableness (A), and Conscientiousness (C).
All trait scores were annotated by trained psychological raters. 
The final labels represent the mean ratings across multiple evaluators, which reduces score variance and strengthens annotation reliability.
Textual data was derived from automatic transcription of audio tracks extracted from the videos.

Following \cite{zhang2025assessing}, the dataset is split using joint stratified sampling over age and work experience to preserve balanced distributions.
The final split includes 450 samples for training, 130 for testing, and 64 for validation, approximating a 7:2:1 ratio. 
We formulate the task as a single-label regression problem, where the model predicts continuous scores for each personality trait independently.

\subsection{Implementation details}
For keyframe extraction, we use a temporal subsampling interval of 10 frames and a Hanning smoothing window of size 50.  
We further analyze the statistics of the keyframe extraction process.  
On average, our method identifies 4.13 keyframes per video (std = 1.79), with a minimum of 0 and a maximum of 12 across the dataset.  
Similar statistics are observed across different subsets, including q3--q6 and training/validation/testing splits, indicating stable behavior of the proposed keyframe detection strategy.

During the AU selection stage, we employed a lightweight LSTM (hidden size = 64, single layer, dropout = 0.1) to jointly model multiple AUs, using validation MSE as the energy function.  
The LSTM was trained for 5 steps per iteration with early stopping (patience = 2).  
Each simulated annealing (SA) run consisted of 50 iterations with a cooling rate of 0.95, where candidate subsets are perturbed via random AU toggling.  
To ensure thorough exploration, three initialization strategies were employed per seed: full AUs subset, single random AU, and random-sized AUs subset. 
Multiple seeds were traversed to collect optimal AUs subsets for subsequent multi-objective optimization.

For AU semanticization, we use GPT-4o (gpt-4o-2024-11-20) via the OpenAI API to generate natural language descriptions for small temporal windows and synthesize them across larger intervals using the structured prompts outlined in Section~\ref{sec:Methodology}.
Temperature and top-p are not explicitly set and therefore follow the default decoding configuration of the API. The seed parameter is not used.
For personality trait assessment, AU-derived semantic descriptions and participants' textual responses were formatted into structured prompts and fed into Meta-Llama-3.1-8B-Instruct.  
The hidden state of the last token was extracted as a semantic embedding and passed to a single-layer regression head.

The LLM was fine-tuned with LoRA, updating low-rank adaptation matrices within the query and value projections (rank = 8, $\alpha$ = 32, dropout = 0.1). The regression head was trained jointly using MSE loss.  
To reduce memory consumption, 4-bit quantization was applied, and gradient accumulation simulated an effective batch size of 8 (batchsize = 2, accumulate steps = 4).  
Training proceeded for up to 100 epochs with early stopping (patience = 5) on vGPU-48GB, employing AdamW with a learning rate of 3e-4 and a cosine scheduler with linear warmup.

\subsection{Evaluation metrics}
We adopt Mean Squared Error (MSE) and Mean Absolute Error (MAE) to evaluate single-label regression performance.  
MAE measures the average magnitude of prediction errors, providing a straightforward interpretation of accuracy: 
\begin{equation}
\text{MAE}_p = \frac{1}{N}\sum_{i=1}^N |y_{i,p} - \hat{y}_{i,p}|,
\label{equ:MAE}
\end{equation}
where \(N\) is the number of samples, \(y_{i,p}\) is the ground-truth score for trait \(p\) of sample \(i\), and \(\hat{y}_{i,p}\) is the predicted score.

Mean Squared Error (MSE) penalizes larger deviations more heavily and reflects overall prediction stability:
\begin{equation}
\text{MSE}_p = \frac{1}{N}\sum_{i=1}^N (y_{i,p} - \hat{y}_{i,p})^2.
\label{equ:MSE}
\end{equation}

Additionally, we report the Pearson correlation coefficient $r$ to assess the linear relationship between predicted and ground-truth scores for each personality trait $p$:
\begin{equation}
r_p = \frac{\sum_{i=1}^N (y_{i,p} - \bar{y}_p)(\hat{y}_{i,p} - \bar{\hat{y}}_p)}
{\sqrt{\sum_{i=1}^N (y_{i,p} - \bar{y}_p)^2} \sqrt{\sum_{i=1}^N (\hat{y}_{i,p} - \bar{\hat{y}}_p)^2}},
\label{equ:pearson}
\end{equation}
where $\bar{y}_p$ and $\bar{\hat{y}}_p$ are the mean values of the ground-truth and predicted scores, respectively. 
We also report the associated significance $p$-value to evaluate whether the observed correlation is statistically different from zero.

\section{Results}
\label{sec:Results}
To evaluate the effectiveness of the proposed AU-text multimodal framework, we conducted a series of experiments on the AVI-6 dataset.  
Tables~\ref{tab:comparison} and~\ref{tab:ablation} summarize the results of the comparison with baselines and ablation studies, respectively. The AVI-6 dataset contains, for each participant, a set of video files corresponding to different personality traits.
It is well-suited for single-label regression tasks.
Our proposed method achieved mean squared errors (MSE) of 0.1555, 0.1380, 0.1597, and 0.1077 on the \textit{Honesty-Humility} (H), \textit{eXtraversion}(X), \textit{Agreeableness}(A), and \textit{Conscientiousness}(C) traits, respectively.  
These relatively low MSE values demonstrate the effectiveness of integrating AU-based semantic information with textual responses for personality assessment.  
Despite relying on only a small number of frames extracted from each video, our framework achieves competitive performance compared to text-only or AU-only baselines.
These results highlight the value of leveraging semantically enriched non-verbal cues to complement linguistic information.

\subsection{Comparison with baselines}
The primary goal of this experiment is not to surpass all baselines, but to evaluate whether integrating AU-based semantic representations with textual responses improves personality trait assessments. 
Due to the lack of directly comparable state-of-the-art methods under the same task setting, we constructed three categories of representative baselines reflecting distinct modeling paradigms.

(1) \textbf{Single-modality baselines} evaluate the predictive capability of each modality independently.
We consider two input-level video representations under a unified evaluation protocol. 
The first is the small window (SW) representation, where each sample consists of a fixed-length temporal segment centered at a keyframe, defined as a window of seven consecutive frames (three preceding and three following the keyframe). All SW segments extracted from a video are ordered chronologically and concatenated to preserve local temporal continuity. 
The second is dense sampling (DS), where frames are uniformly sampled over the entire video sequence to capture global temporal structure.

For video-based baselines, we evaluate ResNet \cite{ResNet} and ViT \cite{ViT}.
ResNet is evaluated under both SW and DS settings, while ViT is applied only under the SW setting due to its computational cost on long sequences. 
The text-only baseline (Longformer \cite{Longformer}) processes only textual responses with instruction components removed to match encoder-only settings. 
For AU-based modeling, we use an LSTM \cite{LSTM} trained on a low-dimensional AU subset selected via simulated annealing to avoid instability caused by high-dimensional raw AU features.

(2) \textbf{Multimodal fusion baselines} combine visual and textual embeddings using classical deep learning strategies.
We adopt classical feature-level and decision-level fusion pipelines using ResNet and BERT \cite{BERT}, 
as well as a state-of-the-art multimodal LLM (Qwen-3-VL-8B-Instruct \cite{Qwen3}) operating in a zero-shot setting to assess the performance of pretrained multimodal alignment without fine-tuning.

Both feature-level and decision-level fusion baselines use SW frames for video and only textual responses for text input.
For the zero-shot multimodal LLM (Qwen-3-VL-8B-Instruct), the same prompt structure is used as in our method while AU semantic descriptions are replaced with video-derived features.

(3) \textbf{AU semantic + text baselines} assess the effect of AU-derived semantic description combined with textual responses.
This includes an encoder-only Longformer that processes AU semantics and interview responses in a unified textual format. This baseline provides a direct comparison for our method under identical input conditions.

The following setup was applied consistently across all baselines to ensure comparability.
All models adhered to the same experimental protocol. 
We used identical data splits, keyframe extraction methods, and standardization steps. A unified linear regression head and fixed random seeds ensured full reproducibility.

\begin{table*}[h!]
\centering
\caption{Comparison with representative baselines. Best results are in \textbf{bold}.}
\label{tab:comparison}
\resizebox{0.9\linewidth}{!}{
\begin{tabular}{llccccc}
\toprule
\multicolumn{7}{c}{Mean Squared Error (MSE)} \\
\midrule
Category & Approaches & H & X & A & C & Avg \\
\midrule
\multirow{5}{*}{Single-modality baselines}
& LSTM (Selected AUs) & 0.2057 & 0.2990 & 0.2188 & 0.2430 & 0.2416 \\
& ResNet (SW) & 0.4583 & 1.3539 & 0.9343 & 0.4982 & 0.8112 \\
& ResNet (DS) & 0.3743 & 0.5765 & 0.4058 & 0.5048 & 0.4654 \\
& ViT (SW) & 0.2489 & 0.3551 & 0.2882 & 0.2785 & 0.2927 \\
& Longformer (only text) & 0.1815 & 0.2697 & 0.2053 & 0.2097 & 0.2166 \\
\midrule
\multirow{3}{*}{Multimodal fusion baselines}
& ResNet + BERT (feature fusion) & 0.2560 & 0.2971 & 0.5280 & 0.3146 & 0.3489 \\
& ResNet + BERT (decision fusion) & 0.6673 & 0.5503 & 0.4162 & 0.4447 & 0.5196 \\
& Qwen-3-VL-8B-Instruct & 1.7777 & 0.4847 & 1.1193 & 1.1653 & 1.1368 \\
\midrule
\multirow{2}{*}{AU semantic + text fusion}
& Longformer (AU semantic + text) & 0.2015 & 0.3050 & 0.2058 & 0.2098 & 0.2305 \\
& Ours & \textbf{0.1555} & \textbf{0.1380} & \textbf{0.1597} & \textbf{0.1077} & \textbf{0.1402} \\
\midrule\midrule
\multicolumn{7}{c}{Mean Absolute Error (MAE)} \\
\midrule
Category & Approaches & H & X & A & C & Avg \\
\midrule
\multirow{5}{*}{Single-modality baselines}
& LSTM (Selected AUs) & 0.3664 & 0.4286 & 0.3858 & 0.3985 & 0.3948 \\
& ResNet (SW) & 0.5436 & 0.9470 & 0.8116 & 0.5777 & 0.7198 \\
& ResNet (DS) & 0.4741 & 0.5987 & 0.5239 & 0.5823 & 0.5448 \\
& ViT (SW) & 0.4084 & 0.4751 & 0.4145 & 0.4203 & 0.4296 \\
& Longformer (only text) & 0.3508 & 0.4009 & 0.3721 & 0.3646 & 0.3721 \\
\midrule
\multirow{3}{*}{Multimodal fusion baselines}
& ResNet + BERT (feature fusion) & 0.4036 & 0.4444 & 0.5769 & 0.4507 & 0.4689 \\
& ResNet + BERT (decision fusion) & 0.6749 & 0.6139 & 0.5166 & 0.5272 & 0.5831 \\
& Qwen-3-VL-8B-Instruct & 1.2245 & 0.5349 & 0.9259 & 0.9574 & 0.9107 \\
\midrule
\multirow{2}{*}{AU semantic + text fusion}
& Longformer (AU semantic + text) & 0.3632 & 0.4183 & 0.3708 & 0.3613 & 0.3784 \\
& Ours & \textbf{0.3242} & \textbf{0.2833} & \textbf{0.3287} & \textbf{0.2565} & \textbf{0.2982} \\
\midrule\midrule
\multicolumn{7}{c}{Pearson Correlation Coefficient (Corr)} \\
\midrule
Category & Approaches & H & X & A & C & Avg \\
\midrule
\multirow{5}{*}{Single-modality baselines}
& LSTM (Selected AUs) & -0.0448 & 0.2236$^{**}$ & -0.0343 & 0.1633 & 0.0770 \\
& ResNet (SW) & 0.0387 & -0.0400 & 0.3675$^{***}$ & -0.0320 & 0.0836 \\
& ResNet (DS) & 0.2014$^{*}$ & -0.1573 & 0.1789$^{*}$ & -0.0235 & 0.0499 \\
& ViT (SW) & 0.1597 & -0.0410 & -0.0131 & 0.1505 & 0.0640 \\
& Longformer (only text) & 0.3279$^{**}$ & 0.2736$^{**}$ & 0.0617 & 0.0022 & 0.1664 \\
\midrule
\multirow{3}{*}{Multimodal fusion baselines}
& ResNet + BERT (feature fusion) & 0.1507 & 0.2573$^{**}$ & 0.1307 & 0.2537$^{**}$ & 0.1981 \\
& ResNet + BERT (decision fusion) & 0.0215 & 0.1065 & 0.0957 & 0.2397$^{**}$ & 0.1159 \\
& Qwen-3-VL-8B-Instruct & 0.3814$^{***}$ & 0.6683$^{***}$ & \textbf{0.5589$^{***}$} & 0.5779$^{***}$ & 0.5467 \\
\midrule
\multirow{2}{*}{AU semantic + text fusion}
& Longformer (AU semantic + text) & 0.1200 & 0.0196 & 0.0199 & 0.0533 & 0.0532 \\
& Ours & \textbf{0.4654$^{***}$} & \textbf{0.7258$^{***}$} & 0.5154$^{***}$ & \textbf{0.6934$^{***}$} & \textbf{0.6000} \\
\bottomrule
\end{tabular}
}
\vspace{0.5em}
\begin{flushleft}
\footnotesize
\textbf{Note:} * $p<0.05$, ** $p<0.01$, *** $p<0.001$.
\end{flushleft}
\end{table*}

Table~\ref{tab:comparison} summarizes the comparison results for the four HEXACO traits in the AVI-6 dataset. As shown, our approach achieves the best performance on nearly all four personality traits (H, X, A, and C) and evaluation metrics. For both MSE and MAE, our model surpasses not only single-modality baselines (AU-numeric, video-only, and text-only) but also multimodal models including the MLLM baseline and traditional fusion strategies. 
In particular, compared with the strongest unimodal baseline, Longformer (only text), our approach yields substantial MSE reductions of 14-49\% depending on the trait.
These results indicate that text features alone do not capture the full range of personality-relevant cues. 
The improvement over the AU-numeric LSTM baseline is even more pronounced, with relative MSE reductions exceeding 25-60\%. 
These consistent gains across two complementary regression metrics highlight the robustness and enhanced predictive precision achieved through our AU semantic-text fusion framework.
Correlation results further validate the effectiveness of our method from a statistical association perspective. 
While most baselines exhibit weak or inconsistent correlations, our approach achieves strong positive correlations for all traits, with coefficients ranging from 0.465 to 0.726. 
All results reach statistical significance at the $p<0.001$ level, indicating a reliable alignment between predictions and personality scores rated by human observers. 
Although the MLLM baseline (Qwen-3-VL-8B-Instruct) also leverages multimodal inputs, its regression accuracy and correlation remain clearly inferior. 
This suggests that multimodal modeling without explicit semantic alignment of AU information yields limited benefit for personality assessment.

In summary, these results provide compelling evidence that our proposed approach enhances regression accuracy. 
It also ensures meaningful and statistically robust correspondence between predictions and human-rated scores, and outperforms all considered baselines across multiple metrics.

\subsection{Ablation experiment}
To quantify the contribution of each component in our method, we conducted ablation experiments by systematically removing individual modules. 
Specifically, we evaluated the following three ablations: 
(1) \textbf{without AU selection}, in which all AUs are used for semantic description without any filtering; 
(2) \textbf{without AU-derived semantic representations}, where only the participants' textual responses are used as input for the personality prediction stage; and 
(3) \textbf{without textual information}, in which only the AU-based semantic descriptions are provided to the personality prediction stage.
\begin{table}[h!]
\centering
\caption{Ablation study results. Lower MSE/MAE indicate better performance. Best results are in \textbf{bold}.}
\label{tab:ablation}
\resizebox{0.95\linewidth}{!}{
\begin{tabular}{lccccl}
\hline
\multicolumn{6}{c}{Mean Squared Error (MSE)}                                                  \\ \hline
Approach           & H & X    & A   & C & Average  \\
w/o AU Selection   & 0.1642           & 0.1695          & 0.2095          & 0.1559            & 0.1748  \\
w/o AU Description & 0.1657           & 0.1613          & \textbf{0.1390} & 0.1627            & 0.1572  \\
w/o Text           & 0.2024           & 0.3210          & 0.2760          & 0.2159            & 0.2538  \\
Ours               & \textbf{0.1555}  & \textbf{0.1380} & 0.1597          & \textbf{0.1077}   & \textbf{0.1402}  \\ \hline
\multicolumn{6}{c}{Mean Absolute Error (MAE)}                                                 \\
w/o AU Selection   & 0.3324           & 0.3097          & 0.3644          & 0.3038            & 0.3276  \\
w/o AU Description & 0.3350           & 0.3120          & \textbf{0.3059} & 0.3094            & 0.3156  \\
w/o Text           & 0.3656           & 0.4382          & 0.4421          & 0.3679            & 0.4035  \\
Ours               & \textbf{0.3242}  & \textbf{0.2833} & 0.3287          & \textbf{0.2565}   & \textbf{0.2982}  \\ \hline
\end{tabular}
}
\end{table}
Table~\ref{tab:ablation} presents the results of our ablation study, in which we systematically assess the contributions of different components of our model: AU description, AU selection, and textual information. 
Performance is reported using Mean Squared Error (MSE) and Mean Absolute Error (MAE), with lower values indicating better predictive accuracy.

The full model achieves the best overall performance across most traits, which confirms the effectiveness of integrating all proposed components.
After removing AU descriptions, the prediction error increases for eXtraversion and Conscientiousness. 
This result suggests that AU-derived semantic representations capture subtle and trait-relevant facial cues that the model would otherwise miss.

When AU selection is omitted, MSE and MAE increase across traits. The degradation is most evident for Agreeableness. 
Thus, the model benefits from retaining only the most informative AUs and from reducing noise introduced by irrelevant facial movements.

The absence of textual information produces the most severe decline in performance. 
Both MSE and MAE rise sharply across all traits. 
For eXtraversion and Agreeableness, the error nearly doubles. 
The ablation results for the text modality align with the nature of the AVI setting, where the interview questions are designed based on trait activation theory. 
Since AVI questions explicitly elicit trait-relevant self-descriptions, the textual responses carry the primary informational content.
Textual responses therefore provide crucial personality evidence while visual cues alone remain insufficient.
Overall, the ablation study confirms that each component (i.e., AU description, AU selection, and text) contributes meaningfully to the model's assessment performance. 
Their combination in the full model yields the most reliable and consistent results.

\subsection{Effect of Temporal Window Size}
We evaluate the effect of different temporal window sizes $N \in \{1,3,5,7,9\}$.
As shown in Table~\ref{tab:window_size}, the performance first improves as the window size increases from 1 to 7, and then degrades when further increasing to 9.
Using a single frame (N=1) leads to inferior and unstable performance. 
More importantly, in the absence of temporal context, the generated descriptions often contain hallucinated temporal dynamics that are not grounded in the input. This observation indicates that single-frame representations are inadequate for dynamic semantic modeling.
Increasing the window size to 3--5 frames provides partial temporal context and leads to improved performance when compared with using a single frame. However, the results remain relatively unstable across different traits, suggesting that such short windows are still insufficient to capture complete facial dynamics.
A window size of 7 nearly achieves the most consistent and best overall performance across all traits.It reflects a suitable balance between temporal context and semantic consistency.
In contrast, larger windows (e.g., 9 frames) introduce noticeable performance degradation (particularly for Conscientiousness), likely due to the inclusion of multiple facial events and increased semantic ambiguity.
Overall, these results suggest that a moderate temporal window is crucial for capturing meaningful facial dynamics while avoiding noise accumulation and semantic mixing.

\begin{table}[h!]
\centering
\caption{Effect of Window Size on Personality Prediction Performance}
\label{tab:window_size}
\resizebox{1\linewidth}{!}{
\begin{tabular}{cccccc}
\hline
\multicolumn{6}{c}{Mean Squared Error (MSE)}                                  \\ \hline
Window Size & H & X & A & C & Average \\
1  & 0.1655          & 0.1444          & 0.1723          & 0.1398          & 0.1555 \\
3  & 0.1646          & 0.1637          & 0.1584          & 0.1209          & 0.1519 \\
5  & 0.1579          & 0.1591          & \textbf{0.1304} & 0.1335          & 0.1452 \\
7  & \textbf{0.1555} & \textbf{0.1380} & 0.1597          & \textbf{0.1077} & \textbf{0.1402} \\
9  & 0.1631          & 0.1418          & 0.1368          & 0.2202          & 0.1655 \\ \hline
\multicolumn{6}{c}{Mean Absolute Error (MAE)}                                  \\ \hline
1  & 0.3334          & 0.2905          & 0.3289          & 0.2815          & 0.3086 \\
3  & 0.3355          & 0.3115          & 0.3249          & 0.2658          & 0.3094 \\
5  & 0.3245          & 0.3082          & \textbf{0.2872} & 0.2826          & 0.3006 \\
7  & \textbf{0.3242} & \textbf{0.2833} & 0.3287          & \textbf{0.2565} & \textbf{0.2982} \\
9  & 0.3340          & 0.2891          & 0.3033          & 0.3694          & 0.3240 \\ \hline
\end{tabular}
}
\end{table}

\subsection{Analysis of AU Subset Selection Strategy}
\begin{figure}[htbp]
    \centering
    \includegraphics[width=\linewidth]{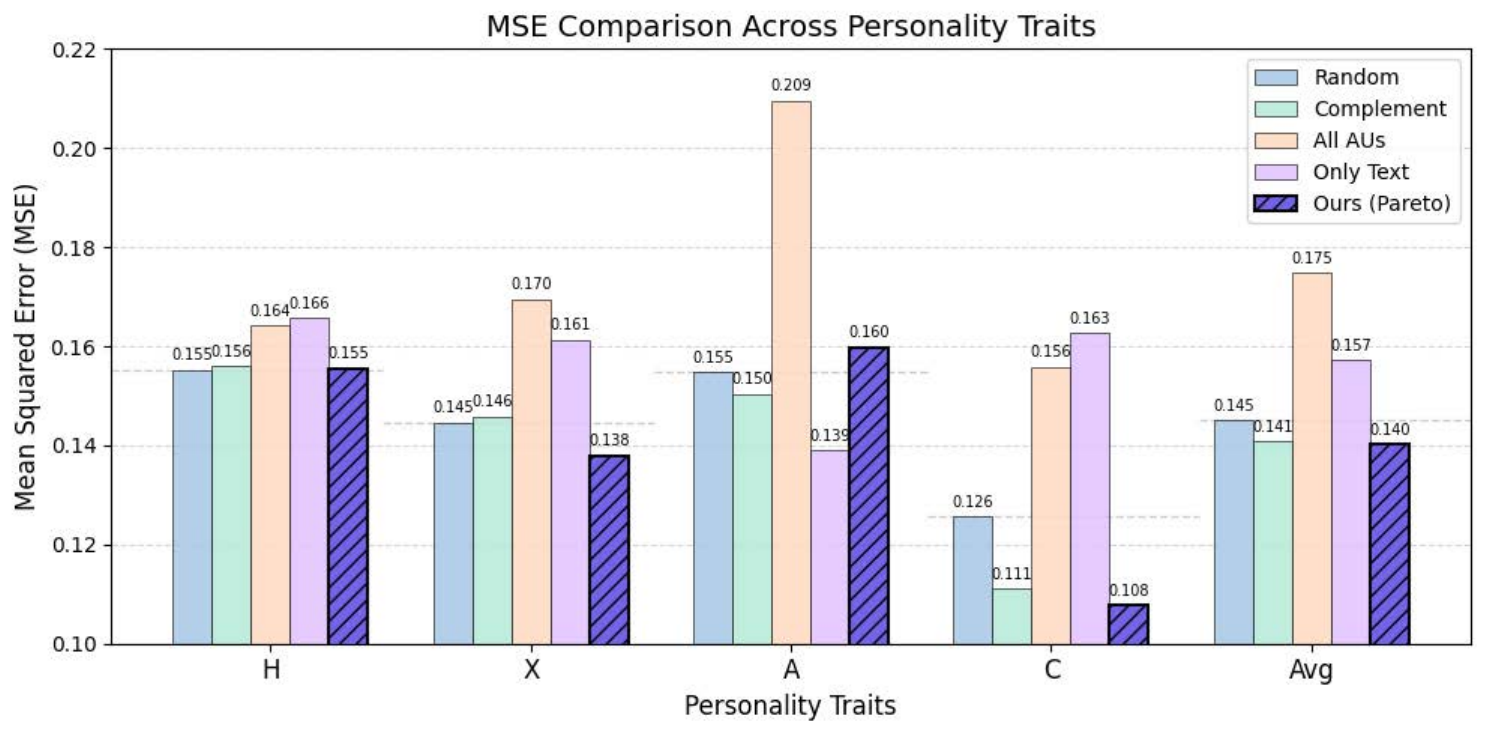}
    \caption{MSE Comparison Across Personality Traits}
    \label{fig:mse_multiple}
\end{figure}

To evaluate the effectiveness of the proposed selection strategy, we compare the Pareto-selected AU subsets with randomly sampled subsets of equal size and complementary subsets.
The results (see Figure~\ref{fig:mse_multiple}) show that the proposed subsets consistently achieve more robust overall performance across traits.
This demonstrates that the selection strategy provides meaningful benefits beyond random or naive alternatives.
Overall, while performance varies across individual traits (e.g., slightly weaker results on Agreeableness), the proposed method achieves slightly better average performance across all metrics.

To further examine whether AU subset selection is non-trivial, we analyze alternative subsets obtained during the simulated annealing optimization process. 
Even under the same temperature configuration, different candidate subsets lead to noticeably different losses. 
This observation confirms that AU selection is not arbitrary and that the search process identifies subsets with distinct behavioral signal characteristics.
In addition, we evaluate the complement subsets formed by the remaining AUs outside the selected Pareto sets. 
These complementary subsets generally yield inferior or less stable performance across multiple traits, further supporting that the selected AUs capture more stable and temporally coherent behavioral signals.

Taken together, these results provide empirical evidence that AU subset selection in the numerical feature space can generalize to the semantic modeling stage.
Although the optimization is performed on raw AU intensity sequences, the selected subsets preserve temporally coherent behavioral patterns that remain meaningful after semantic transformation.
This can be explained by the fact that AU-derived semantic descriptions are constructed from underlying facial movement dynamics.
Therefore, subsets that capture stable and consistent behavioral patterns in the numerical domain are more likely to yield structured and informative semantic representations, which can be effectively exploited by the LLM for personality inference.

\begin{figure}[htbp]
    \centering
    \includegraphics[width=\linewidth]{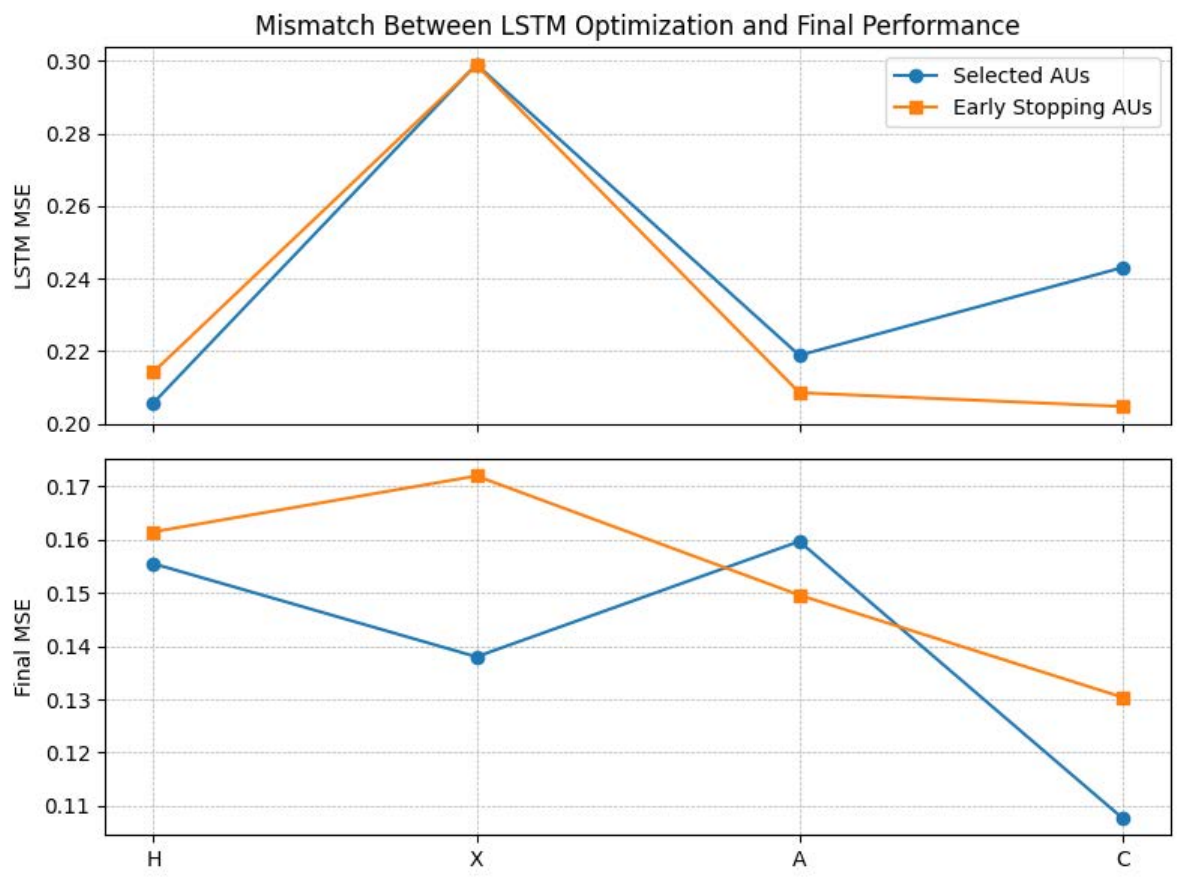}
    \caption{Mismatch Between LSTM Optimization and Final Performance}
    \label{fig:mismatch}
\end{figure}

To further analyze the behavior of the LSTM-based objective, we we evaluate the relationship between LSTM-based energy and final personality prediction performance across different AU subsets (see Fig~\ref{fig:mismatch}).
We observe that subsets with lower LSTM-based MSE do not consistently achieve better final performance. 
For instance, for traits eXtraversion and Conscientiousness, early-stage subsets obtained during simulated annealing exhibit comparable or even lower LSTM MSE, yet result in noticeably worse final prediction accuracy compared to the Pareto-selected subsets.
This indicates that the LSTM-based objective captures certain aspects of temporal structure in AU sequences, but its alignment with downstream semantic-level prediction performance is limited.
At the same time, we observe that subsets with extremely high LSTM-based MSE tend to exhibit degraded downstream performance.
It reflects that the objective is still useful in identifying temporally inconsistent representations.
Overall, these results indicate a partial but not complete correspondence between the surrogate temporal objective and downstream personality prediction performance, which also highlights the necessity of additional selection criteria beyond temporal coherence alone.

\section{Discussion}
\subsection{Is AU effective as supplementary information?}
In AVI-based personality assessment, habitual facial behaviors elicited by personality-relevant questions provide rich non-verbal cues for assessment. 
To better understand how these cues contribute to assessment, we analyzed the subsets of AUs automatically selected for each personality trait. 
Table~\ref{tab:selected_AUs} summarizes the selected AUs and their high-level psychological interpretations.

\begin{table*}[h!]
\centering
\caption{Selected AUs for each personality trait and their psychological interpretation.}
\label{tab:selected_AUs}
\resizebox{0.9\linewidth}{!}{
\begin{tabular}{l l l}
\toprule
Personality Trait & Selected AUs & Psychological Interpretation \\
\midrule
Honesty-Humility & AU06, AU09, AU12, AU17, AU20, AU23, AU25, AU45 & Attentiveness, emotional sincerity \\
eXtraversion      & AU02, AU05, AU06, AU07, AU12, AU14, AU17, AU23, AU45 & Enthusiasm, social engagement \\
Agreeableness     & AU05, AU06, AU12, AU14, AU15, AU20, AU26 & Empathy, positive affect \\
Conscientiousness & AU02, AU10, AU12, AU20, AU23, AU26 & Controlled and focused expression \\
\bottomrule
\end{tabular}
}
\end{table*}

Overall, AUs 6, 12, 20, and 23 consistently appeared across multiple traits, which indicates their role as stable cross-trait signals.
Prior affective computing studies have reported that AU12, AU06, and AU45 are among the most frequently associated facial action units with expressive intensity and social signaling behaviors \cite{10428080}.
This result suggests their dominant role in conveying affective information in spontaneous facial behavior.
Within the same line of research, AU12 in particular has been observed to exhibit relatively stable correlations with personality-related dimensions across multiple datasets (e.g. ChaLearn 2016 and UDIVA), where AU-level facial activity demonstrates non-trivial CCC associations with Big Five traits \cite{10428080}.

Psychologically, these AUs reflect affective and social signaling processes that are relevant across personality traits. 
AU06 (cheek raiser) and AU12 (lip corner puller) together constitute the Duchenne smile, a marker of genuine positive affect and affiliative intent, which is relevant for traits such as Honesty-Humility (sincerity) and Extraversion (sociability) \cite{FACS, EkmanPaul1990TDSE}. 
AU20 (lip stretcher) and AU23 (lip tightener) modulate lip shape to convey attentional focus, evaluative judgment, or subtle tension, which can be informative for Conscientiousness (controlled, attentive expression) and Agreeableness (empathic or considerate responses) \cite{BabaeiEbrahim2020FoFA}.

Their cross-dimensional presence can be partly explained by overlapping muscle activations \cite{FACS, Wingenbach2022_EMG}. 
AU06 and AU12 share engagement of the zygomaticus major and orbicularis oculi muscles. 
AU20 and AU23 involve perioral muscles that adjust lip tension and width. 
These shared muscular pathways enable AUs to convey nuanced affective states across traits, making them stable and generalizable indicators in personality assessment \cite{AU16PF}.

At the same time, prior work has also reported that certain personality traits, such as Agreeableness, tend to exhibit lower correlations with non-verbal facial cues \cite{10428080}.
This observation is consistent with our experimental findings, where the proposed model performs slightly lower than the text-only baseline on the Agreeableness dimension.
One possible explanation is that Agreeableness tends to manifest more strongly through linguistic and social behaviors rather than transient facial expressions. 

It is important to highlight that the effectiveness of AU-derived semantic representations depends on the model's semantic reasoning capability. 
For example, in our experiments, Longformer (encoder-only) using AU semantic description and text as input performed worse than Longformer using only text. 
This suggests that AU semantics alone do not guarantee performance improvements unless the model has sufficient knowledge and capacity to capture the embedded psychological cues. 
In contrast, our LLM-based framework leverages pre-trained language knowledge to fully exploit AU semantic information.
As a result, our method could get improvements across multiple traits.
The observed alignment suggests that behaviorally meaningful patterns may emerge from data-driven selection without explicit psychological supervision.

Overall, our results indicate that fine-grained facial action information provides valuable non-verbal cues for personality assessment.
It also corroborates the \textit{Brunswik's Lens Model} in psychology: personality is expressed not only through verbal behavior but also through characteristic patterns of facial movement.

\subsection{Why semantic understanding should be decoupled from regression prediction?}
Large language models (LLMs) are inherently designed as next-token predictors, optimized for discrete token generation rather than continuous value (i.e, oating-point scorces such as 4.25) estimation.
Their output space is categorical by nature, where logits are normalized by a softmax function to model token probabilities.
Consequently, when directly fine-tune LLMs to predict continuous personality scores, the model faces a representation mismatch.
Since semantic reasoning in embedding space does not align with regression objectives in Euclidean space \cite{understandingllmembeddingsregression}, directly fine-tuning LLMs with continuous values (e.g., Agreeableness = 4.25) often leads to unstable training dynamics, limited value precision, and reduced generalization.

\begin{table}[h!]
\centering
\caption{Effect of Decoupling Semantic Understanding and Regression in LLM-based Personality assessment}
\label{tab:llmregression}
\resizebox{1\linewidth}{!}{
\begin{tabular}{cccccc}
\hline
\multicolumn{6}{c}{Mean Squared Error (MSE)}                                  \\ \hline
Approach & H & X & A & C & Average \\
Zero-shot & 0.6324           & 0.6296       & 0.5679        & 0.6857            & 0.6289  \\
Ours      & \textbf{0.1555}  & \textbf{0.1380} & \textbf{0.1597}          & \textbf{0.1077}   & \textbf{0.1402}  \\ \hline
\multicolumn{6}{c}{Mean Absolute Error   (MAE)}                               \\
Zero-shot & 0.6493           & 0.6553       & 0.6110        & 0.6838            & 0.6499  \\ 
Ours      & \textbf{0.3242}  & \textbf{0.2833} & \textbf{0.3287}          & \textbf{0.2565}   & \textbf{0.2982}  \\ \hline
\end{tabular}
}
\end{table}

Our empirical results confirm this limitation. 
When the LLM was directly trained to output floating-point scores, performance degraded across all traits, as shown in Table~\ref{tab:llmregression}. The model achieves MSE ranging from 0.5679 (A) to 0.6857 (C), with an overall average MSE of 0.6289. 
Similarly, the MAE span from 0.6110 (A) to 0.6838 (C), averaging 0.6499 across all traits. 
These underperformed results indicate that the model struggled to learn fine-grained numerical mappings from textual prompts. 

\begin{figure}[htbp]
    \centering
    \includegraphics[width=\linewidth]{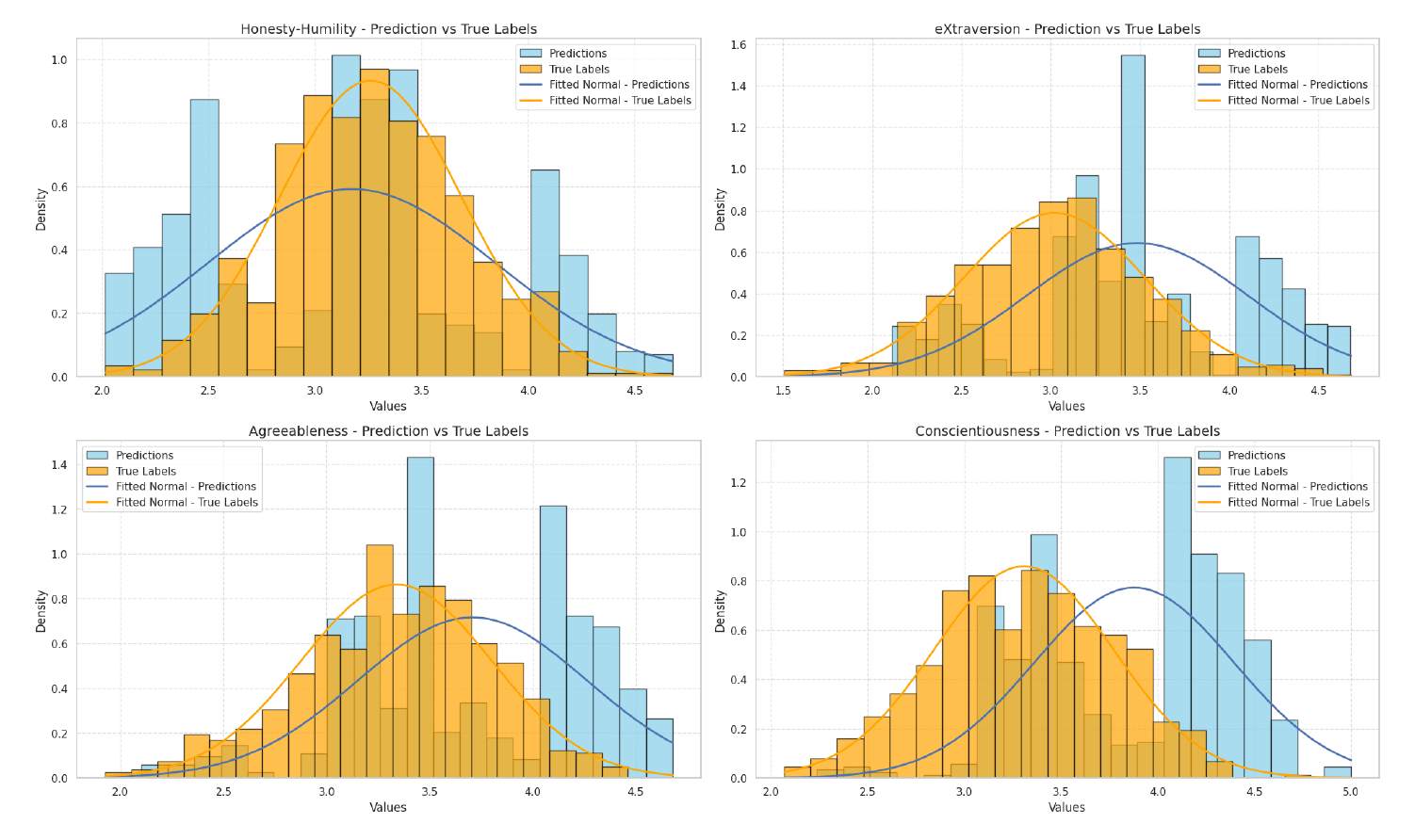}
    \caption{Frequency distributions of predicted personality scores by zero-shot LLM}
    \label{fig:llm_zero_shot}
\end{figure}

Figure~\ref{fig:llm_zero_shot} shows the frequency distribution of the predicted personality scores across the four HEXACO traits. Personality traits are known to exhibit approximate normality, with skewness and kurtosis values typically within conventional bounds (e.g., $|\mathrm{skew}|, |\mathrm{kurtosis}| \lesssim 1.5$) \cite{alma99325356501751, normal_distribution}. 
By contrast, our zero-shot predictions display asymmetric and flattened distributions. 
Three traits show right-skewness, while one exhibits a slight left-skew. 
All predicted distributions have reduced kurtosis relative to empirical norms, indicating that the LLM-generated scores deviate from the expected normal-like patterns and lose the typical concentration around the mean. 
These distributional distortions highlight the model's limited ability to infer fine-grained quantitative differences directly from textual semantics without task-specific calibration.

\begin{figure}[htbp]
    \centering
    \includegraphics[width=\linewidth]{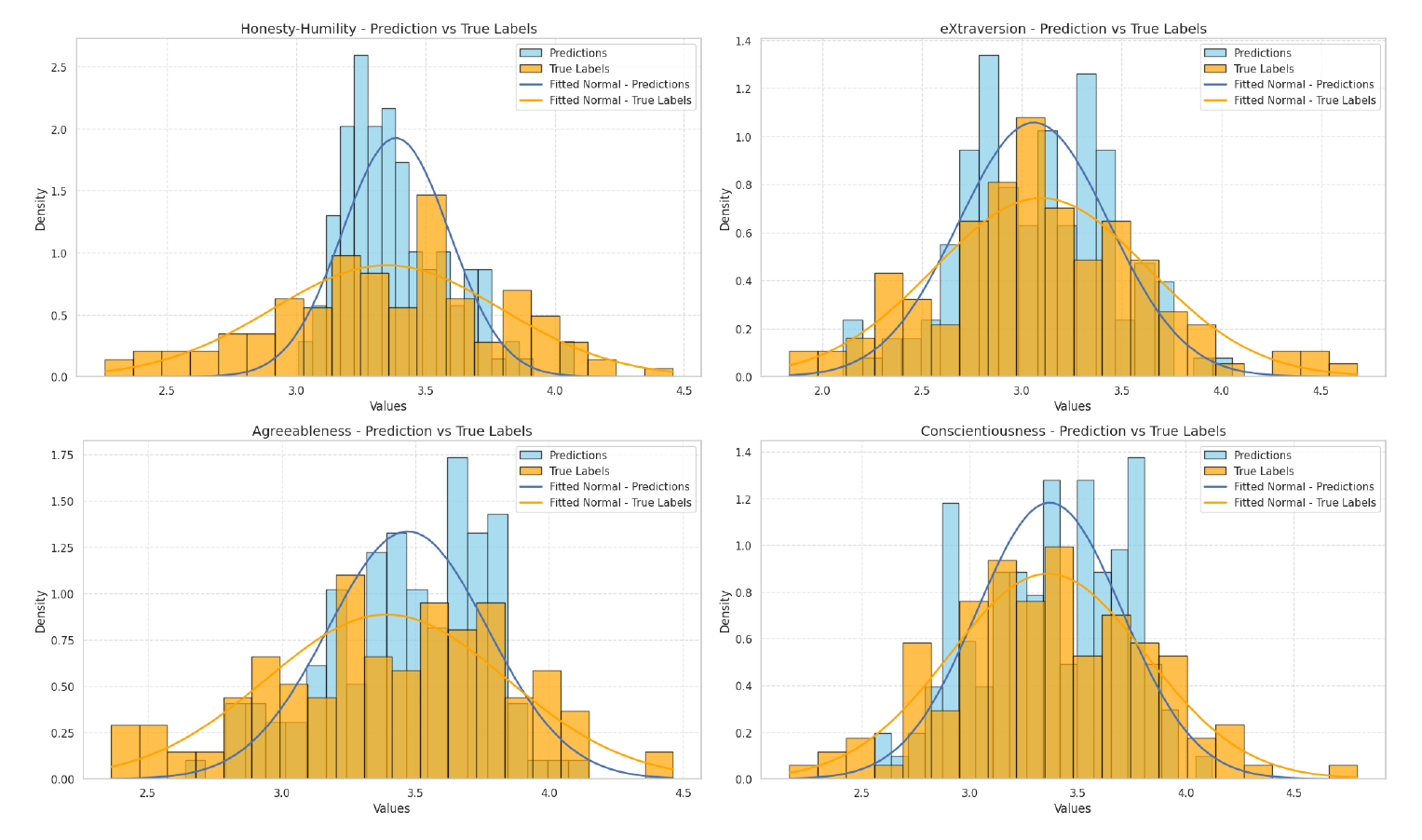}
    \caption{Frequency distributions of predicted personality scores by our method}
    \label{fig:ours_hist}
\end{figure}

In contrast, our decoupled architecture produces more coherent and statistically stable predictions. Figure~\ref{fig:ours_hist} shows that the predicted distributions closely approximate empirical norms. 
They exhibit reduced skewness and stronger concentration around the mean. 
For Honesty-Humility, the distribution becomes more centralized and displays a pronounced peak. 
Extraversion and Conscientiousness present moderate increases in peak height and an improved alignment of the mean with empirical data. 
These results indicate that the regression head preserves the global statistical structure and improves local assessment accuracy. 
Agreeableness remains slightly right-skewed, consistent with its relatively higher MSE compared to the text-only approach. 
This pattern implies that the trait is more sensitive to subtle semantic and behavioral discrepancies.

This improvement arises from the model's design, which explicitly separates semantic encoding from numerical regression.
The LLM focuses on extracting psychologically meaningful representations. The lightweight regression head independently learns the mapping between semantic embeddings and continuous personality scores.
This division of labor reduces optimization interference between linguistic and numerical objectives and promotes smoother convergence.
The LoRA fine-tuning strategy further ensures that the LLM can adapt its semantic space to the downstream regression task without compromising interpretability.

Collectively, these findings suggest that disentangling semantic understanding from regression prediction not only stabilizes learning dynamics but also provide personality traits which are both numerically reliable and semantically coherent.

\section{Limitation and Future Work}

Despite achieving competitive results across multiple personality trait assessment tasks, our approach has several limitations. 
First, variability in video quality, such as poor lighting or incomplete facial visibility, increases the difficulty of AU extraction and poses challenges for subsequent AU semantic processing. 
In addition, although we employ structured prompts and manual calibration, the LLM may still produce information unrelated to the intended AU semantics, reducing the density of task-relevant cues for downstream fusion. 
Second, resource and architectural constraints limit the integration of AU selection with downstream assessment. 
The gap between the LSTM architecture and semantic processing further constrains AU subset selection, while large-scale textual inputs limit our ability to explore multi-label personality assessment and inter-trait correlations.

Future work will focus on addressing these limitations. 
We plan to adopt more robust AU extraction tools to obtain high-quality and reliable facial action data. 
Moreover, we aim to explore token compression and end-to-end optimization strategies to integrate AU selection, semantic generation, and personality assessment into a unified framework, thereby enhancing both information utilization and predictive accuracy.

\section{Conclusion}
In this work, we propose a multimodal framework that incorporates facial action units (AUs) as visual cues to enrich textual features for personality recognition. 
AU sequences are converted into natural-language semantic descriptions, which preserve dynamic facial patterns while maintaining computational efficiency. 
These AU-derived semantic descriptions are then fused with textual responses within a unified language modeling framework. 
A lightweight regression head maps the fused representations into continuous personality trait scores.

Compared with traditional AU-only or text-only approaches, the proposed method delivers superior predictive accuracy, enhanced stability, and improved interpretability. 
The results confirm that explicit semantic alignment between facial dynamics and linguistic content allows the model to capture psychologically meaningful cues that are otherwise lost in either modality alone.

Overall, this framework strengthens the connection between nonverbal behavior and verbal communication, and establishes a scalable pathway for robust and interpretable multimodal personality assessment grounded in psychological constructs.

\bibliography{IEEEabrv, references}

\clearpage
\onecolumn   

\section*{Supplementary Material}

\setcounter{section}{0}
\renewcommand{\thesection}{\Alph{section}}

\newtheorem{theorem}{Theorem}
\newtheorem{definition}{Definition}
\newtheorem{lemma}{Lemma}


\section{AU Semantic Definition}
This work employs a fixed set of 17 Facial Action Units (AUs) as a semantic reference for all AU-based textual descriptions. Each AU is associated with a standardized natural language definition used consistently across all experiments. 

Among them, 14 AU definitions are adopted from \cite{yang2021exploiting}, and the remaining three (AU5, AU20, AU45) follow the Facial Action Coding System (FACS) guidelines \cite{ekman2002facs}. 

All definitions are formatted into a unified descriptive style to ensure consistency in presentation. The complete set is provided in Table~\ref{tab:au_definitions}.

\begin{table*}[htbp]
\centering
\caption{Semantic definitions of 17 Facial Action Units (AUs).}
\label{tab:au_definitions}
\small
\begin{tabular}{p{0.08\linewidth} p{0.88\linewidth}}
\hline
\textbf{AU} & \textbf{Semantic Definition} \\
\hline
AU01 & The inner corners of the eyebrows are lifted slightly, the skin of the glabella and forehead above it is lifted slightly and wrinkles deepen slightly and a trace of new ones form in the center of the forehead. \\
AU02 & The outer part of the eyebrow raise is pronounced. The wrinkling above the right outer eyebrow has increased markedly, and the wrinkling on the left is pronounced. \\
AU04 & Vertical wrinkles appear in the glabella and the eyebrows are pulled together. The inner parts of the eyebrows are pulled down a trace on the right and slightly on the left with traces of wrinkling at the corners. \\
AU05 & The upper lip is raised vertically, exposing more of the teeth and gums slightly. The skin above the lip stretches upward, creating subtle wrinkles or tension lines between the nose and the upper lip. \\
AU06 & The cheeks are lifted without raising the lip corners. The infraorbital furrow has deepened slightly, and wrinkles under the eyes increase. \\
AU07 & The lower eyelid is raised markedly, causing bulging and narrowing of the eye aperture. \\
AU09 & Wrinkling the nose and lifting the nasal wings, which deepens the upper nasolabial fold as the upper lip is drawn up. \\
AU10 & The upper lip is drawn straight up, the outer portions of the lip are raised slightly. \\
AU12 & The corners of the lips are raised obliquely. \\
AU14 & The lip corners are tightly pulled inward, causing significant wrinkling around the corners and stretching of the skin on the chin and lower lip. \\
AU15 & The corners of the lips are pulled down slightly, with some lateral pulling. \\
AU17 & Severe wrinkling of the chin with the lower lip pushed up and out. \\
AU20 & The corners of the lips are pulled outward horizontally, the skin around the mouth is stretched thinly, and there is a noticeable tension along the line extending from the corners of the lips. \\
AU23 & The lips are tightly pressed, narrowing the red parts and causing significant wrinkling around the lips. \\
AU25 & The teeth are visible with lips slightly parted. \\
AU26 & The jaw drops as much as possible, parting the lips. \\
AU45 & The eyelids close abruptly for a brief moment, with the upper eyelid dropping sharply and the lower eyelid lifting slightly, causing the eye aperture to disappear temporarily before quickly reopening. \\
\hline
\end{tabular}
\end{table*}

This definition set is kept fixed across all experiments and is not modified during training or inference.

\section{Full Prompt Design}

To ensure reproducibility, we provide the full prompts used in our pipeline. 
All prompts are presented in their original form with minor formatting adjustments 
for readability. The content is identical to the implementation used in our experiments.


\subsection{Prompt 1: Semantic descriptions within small windows generation}

This prompt converts frame-level AU intensity signals into natural language descriptions.

\begin{tcolorbox}

\textbf{Role Definition:}\\
You are currently acting as an expert in describing facial action dynamics 
based on Action Unit (AU) data.

\vspace{0.5em}
\textbf{Task:}\\
Your goal is to generate a succinct, natural-language summary of the facial 
movement trends covering the entire duration of the specified AUs.

\vspace{0.5em}
\textbf{AU Semantic Definition:}\\
\{All AU semantic definitions are provided here exactly as listed in Section 1.\}

\vspace{0.5em}
\textbf{Instructions:}
\begin{itemize}
\item Do NOT mention any AU identifiers (e.g., AU06, AU12) or numeric intensity values.
\item Begin your description with the frame range in this exact format: 
  "From frames XX to YY,".
\item Summarize only physical AU activity and observable trends.
\item Avoid referencing AUs not included in the provided list.
\item Write a single continuous paragraph.
\item Do NOT include field names such as "Input:" or "Output:".
\item Use the example only as a style reference.
\item If any AU data is ambiguous, describe only clearly observable patterns.
\end{itemize}

\vspace{0.5em}
\textbf{Example:}\\
\textit{Input (AU Data):}

\begin{center}
\small
\begin{tabular}{c|cccccccc}
\hline
Frame & AU06 & AU09 & AU12 & AU17 & AU20 & AU23 & AU25 & AU45 \\
\hline
146 & 1.84 & 0.05 & 0.04 & 0.00 & 0.00 & 0.06 & 0.71 & 0.25 \\
147 & ...  & ...  & ...  & ...  & ...  & ...  & ...  & ...  \\
148 & ...  & ...  & ...  & ...  & ...  & ...  & ...  & ...  \\
149 & ...  & ...  & ...  & ...  & ...  & ...  & ...  & ...  \\
150 & ...  & ...  & ...  & ...  & ...  & ...  & ...  & ...  \\
151 & ...  & ...  & ...  & ...  & ...  & ...  & ...  & ...  \\
152 & 1.85 & 0.04 & 0.10 & 0.11 & 0.12 & 0.00 & 0.24 & 0.31 \\
\hline
\end{tabular}
\end{center}

\vspace{0.3em}
\textit{Output:}\\
From frames 146 to 152, the cheeks remain gently lifted, sustaining a consistent upward presence throughout the sequence...

\vspace{0.5em}
\textbf{Input (AU Data):}\\
\{Window-specific AU table\}
\end{tcolorbox}

\vspace{0.5em}
\textbf{Note:}
In practice, the subset of AUs and corresponding semantic definitions are 
dynamically selected based on the target trait (q3--q6). This does not change 
the prompt structure but only the provided AU list.


\subsection{Prompt 2: Global summary descriptions generation}

This prompt aggregates multiple short-window descriptions into a coherent 
long-range temporal summary.

\begin{tcolorbox}

\textbf{Role Definition:}\\
You are assisting in analyzing Facial Action Unit (AU) trends from videos.

\vspace{0.5em}
\textbf{Task:}\\
Your task is to integrate AU descriptions from multiple short windows into 
a single cohesive description that reflects the overall temporal trend.

\vspace{0.5em}
\textbf{Instructions:}
\begin{itemize}
\item Do NOT concatenate or copy-paste descriptions.
\item Generate a unified and coherent summary.
\item Avoid redundancy and list-style outputs.
\item Present facial dynamics as a smooth progression.
\item Use a merged frame range (e.g., "From frames 61 to 152").
\item Ignore missing intermediate frames.
\item Focus on temporal evolution and behavioral consistency.
\end{itemize}

\vspace{0.5em}
\textbf{Example:}\\
\textit{Input:}\\
description1: ...\\
From frames 61 to 67, the brows show a steady downward movement...\\
description2: ...\\
From frames 146 to 152, the cheeks exhibit a subtle lift...\\
\textit{Output:}\\
From frames 61 to 152, the facial dynamics reflect gradual and shifting 
patterns across the brow, cheeks, chin, and eyes...

\vspace{0.5em}
\textbf{Input:}\\
description1: ...\\
\{Short-window description\} ...\\
description2: ...\\
\{Short-window description\}
\end{tcolorbox}


\subsection{Prompt 3: Personality Trait Embedding Generation}

This prompt integrates multimodal cues for personality-aware representation learning.

\begin{tcolorbox}

\textbf{Context setup:}\\
The subject was asked to record a video in response to the personality-eliciting question: \{question\_text\}. 

\vspace{0.5em}
\textbf{Multimodal information presentation:}\\
Information extracted from the video:\\
1.Semantic description of facial action units trends: \{description\}\\
2.Textual answers:\{answer\}

\vspace{0.5em}
\textbf{Trait Definition:}
We use standardized HEXACO-PI-R trait definitions from the official HEXACO scale descriptions 
(\url{https://hexaco.org/scaledescriptions}). The definitions are provided verbatim in the implementation 
and dynamically inserted at runtime as \{trait\_definition\}.

\vspace{0.5em}
\textbf{Role Definition:}\\
You are an expert in the HEXACO personality model.

\vspace{0.5em}
\textbf{Task:}\\
Your task is to analyze the {trait} subdimension and generate an embedding that captures personality-relevant patterns from the information and definition.

\end{tcolorbox}

\vspace{0.5em}
\textbf{Note:}
The embedding is generated implicitly by the language model and subsequently 
used for downstream regression without additional manual feature engineering.



\end{document}